\documentclass[runningheads]{llncs}

\usepackage{eccv}

\usepackage{eccvabbrv}

\usepackage{graphicx}
\usepackage{booktabs}

\usepackage[accsupp]{axessibility}  % Improves PDF readability for those with disabilities.

\usepackage[pagebackref,breaklinks,colorlinks]{hyperref}
\usepackage{orcidlink}

\usepackage{titlecaps}
\usepackage{microtype}
\usepackage{pdfpages}

\usepackage{tabularx,ragged2e}
\begin{document}

\newcommand{\bench}{{\textsc{Blink}}\xspace}

\newcommand{\ensuretext}[1]{#1}
\newcommand{\marker}[2]{\ensuremath{^{\textsc{#1}}_{\textsc{#2}}}}
\newcommand{\arkcomment}[3]{\ensuretext{\textcolor{#3}{[#1 #2]}}}
\newcommand{\yushi}[1]{\arkcomment{\marker{Y}{H}}{#1}{red}}
\definecolor{darkgreen}{rgb}{0,0.5,0}
\newcommand{\weichiu}[1]{\arkcomment{\marker{W}{M}}{#1}{red}}
\newcommand{\wc}[1]{\textcolor{blue}{#1}}
\definecolor{pink}{RGB}{245, 66, 141}
\newcommand{\fxy}[1]{\arkcomment{\marker{X}{F}}{#1}{pink}}
\newcommand{\xingyu}[1]{\textcolor{black}{#1}}

\newcommand{\ranjay}[1]{\textcolor{blue}{Ranjay: #1}}
\newcommand{\nascomment}[1]{\textcolor{red}{Noah: #1}}
\newcommand{\todo}{\hl{TODO}\ }
\newcommand{\name}{BLINK}

% ---------------------------------------------------------------
% TODO REVIEW: Replace with your title
% \title{\name: Benchmarking Perceptual Abilities not yet Observed in Multimodal Large Language Models}
\title{\name \includegraphics[scale=0.03]{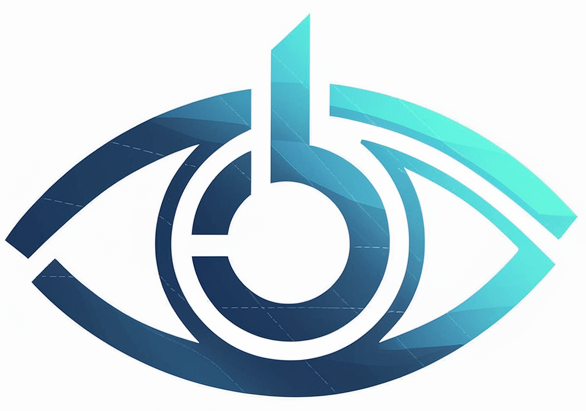}: Multimodal Large Language Models Can See but Not Perceive}
% Blink: The Power of Thinking without Thinking: Limitations of Perception Abilities in Large Language Models
% In the Blink of an Eye: Visual Perception Abilities that have not Emerged yet on Multimodal Large Language Models
% \name： Perception Abilities not Yet Emerged on Multimodal Large Language Models

\titlerunning{\bench}

\author{Xingyu Fu\inst{1}\thanks{Equal contribution. Correspond to <Xingyu Fu: xingyuf2@seas.upenn.edu>, <Yushi Hu: yushihu@uw.edu>. All data and evaluation are available on the project page.},  
Yushi Hu\inst{2,3}$^{*}$,  
Bangzheng Li\inst{4}, 
Yu Feng\inst{1},
Haoyu Wang\inst{1}, 
Xudong Lin\inst{5},
Dan Roth\inst{1},
Noah A. Smith\inst{2,3},
Wei-Chiu Ma\inst{3}\thanks{Both authors advised equally.},
Ranjay Krishna\inst{2,3}$^{\dagger}$\\
\vspace{1ex}
\small{
\inst{1}University of Pennsylvania, 
\inst{2}University of Washington,
\inst{3}Allen Institute for AI,\\
\inst{4}University of California, Davis, 
\inst{5}Columbia University%, 
%\inst{6}Cornell University\\
}
% \texttt{xingyuf2@seas.upenn.edu}
\url{https://zeyofu.github.io/blink/}
}

% % TODO FINAL: Replace with your author list. 
% % Include the authors' OCRID for the camera-ready version, if at all possible.
% \author{First Author\inst{1}\orcidlink{0000-1111-2222-3333} \and
% Second Author\inst{2,3}\orcidlink{1111-2222-3333-4444} \and
% Third Author\inst{3}\orcidlink{2222--3333-4444-5555}}

% TODO FINAL: Replace with an abbreviated list of authors.
\authorrunning{Fu \etal.}
% First names are abbreviated in the running head.
% If there are more than two authors, 'et al.' is used.

% % TODO FINAL: Replace with your institution list.
% \institute{University of Pennsylvania \and University of Washington
% \and University of California, Davis
% \and Columbia University
% \and Cornell University
% \email{\{abc,lncs\}@uni-heidelberg.de}}

\makeatletter
\makeatother

\newcommand{\iiw}{{relative reflectance}\xspace}
\newcommand{\dreamsim}{{visual similarity}\xspace}
\newcommand{\iq}{{IQ test}\xspace}
\newcommand{\realness}{{forensics detection}\xspace}
\newcommand{\funcCorr}{{functional correspondence}\xspace}
\newcommand{\semCorr}{{semantic correspondence}\xspace}
\newcommand{\camerapose}{{multi-view reasoning}\xspace}
\newcommand{\spatial}{{spatial relation}\xspace}
\newcommand{\counting}{{counting}\xspace}
\newcommand{\art}{{art style}\xspace}
\newcommand{\jigsaw}{{jigsaw}\xspace}
\newcommand{\corr}{{visual correspondence}\xspace}
\newcommand{\localization}{{object localization}\xspace}
\newcommand{\depth}{{relative depth}\xspace}

\newcommand{\iiwshort}{{Reflect.}\xspace}
\newcommand{\dreamsimshort}{{Similarity}\xspace}
\newcommand{\iqshort}{{IQ}\xspace}
\newcommand{\realnessshort}{{Forensic}\xspace}
\newcommand{\funcCorrshort}{{Fun.Corr.}\xspace}
\newcommand{\semCorrshort}{{Sem.Corr.}\xspace}
\newcommand{\cameraposeshort}{{Multi-view}\xspace}
\newcommand{\spatialshort}{{Spatial}\xspace}
\newcommand{\countingshort}{{Counting}\xspace}
\newcommand{\artshort}{{Art}\xspace}
\newcommand{\jigsawshort}{{Jigsaw}\xspace}
\newcommand{\corrshort}{{Vis.Corr.}\xspace}
\newcommand{\localizationshort}{{Local.}\xspace}
\newcommand{\depthshort}{{Depth}\xspace}

\maketitle

% \titlehook{
\vspace{-32pt}
\begin{center}
    \centering
    \captionsetup{type=figure}
    \includegraphics[width=0.9\textwidth,trim={0cm 0 0cm 0},clip]{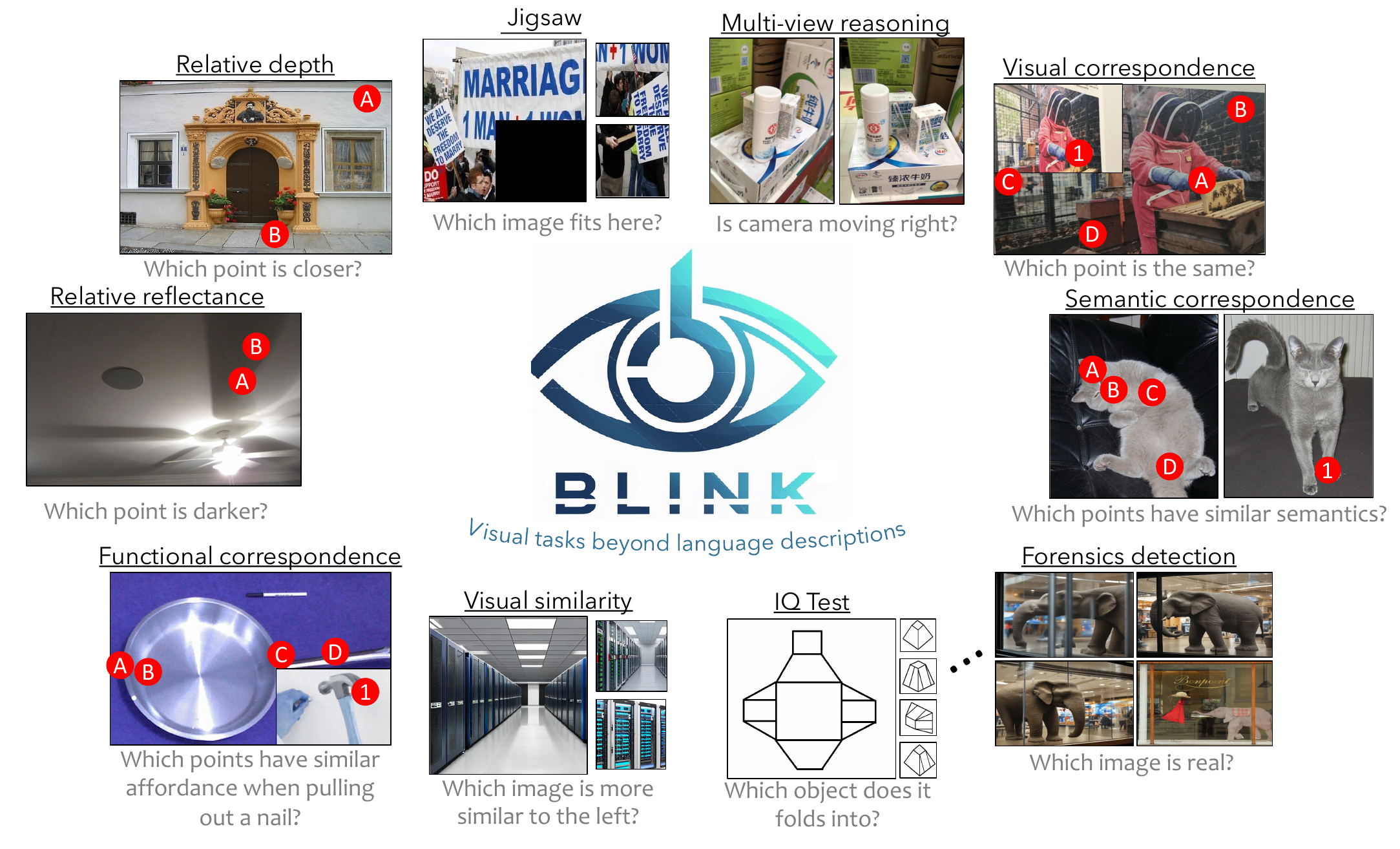}
\captionof{figure}{{\bf The \bench Benchmark.} \bench contains 14 visual perception tasks that can be solved by humans ``within a blink'', but pose significant challenges for current multimodal LLMs. 
These tasks are inspired by classical computer vision problems and recast into multiple-choice questions for multimodal LLMs to answer.
Notice that the visual prompts and questions in this figure are different from the actual ones used in the benchmark for illustrative purposes, and answers of the samples are provided.\footref{explainfootnote}
}
\label{fig:teaser}
    \vspace{-2ex}
    
\end{center}%
% }

\begin{abstract}
    We introduce~\bench, a new benchmark for multimodal language models (LLMs) that focuses on core visual perception abilities not found in other evaluations.
Most of the \bench tasks can be solved by humans ``within a blink'' (\eg, relative depth estimation, visual correspondence, forensics detection, and multi-view reasoning). However, we find these perception-demanding tasks cast significant challenges for current multimodal LLMs because they resist mediation through natural language.
\bench reformats 14 classic computer vision tasks into 3,807 multiple-choice questions, paired with single or multiple images and visual prompting.
% \bench is comprised of 3,978 multiple-choice questions paired with single or multiple images and visual prompting, across 14 tasks meticulously selected to be naturally easy for humans and inspired by classical computer vision perception problems. 
While humans get 95.70\% accuracy on average, \bench is surprisingly challenging for existing multimodal LLMs:
even the best-performing GPT-4V and Gemini achieve accuracies of 51.26\% and 45.72\%, only 13.17\% and 7.63\% higher than random guessing, indicating that such perception abilities have not ``emerged'' yet in recent multimodal LLMs.
% In our evaluation on \bench, we benchmarked 16 different models, highlighting a gap of 40\% accuracy achieved by human over the state-of-the-art GPT-4V, indicating significant room for improvement. 
% Our analysis not only highlights the observed limitations of recent multimodal LLMs but also shows 
% Further, the 7B, 13B, and 34B multimodal LLMs perform similarly, indicating that LM scaling (at least in dimensions explored so far) does not help with \bench tasks. 
Our analysis also highlights
that specialist CV models could solve these problems much better, suggesting potential pathways for future improvements. 
%\bench can also serve as a testbed for visual prompting techniques.
We believe \bench will stimulate the community to help multimodal LLMs catch up with human-level visual perception.
% Project link~\url{https://huggingface.co/PerceptionEval}. \fxy{build page}
 
\end{abstract}

\section{Introduction}
\label{sec:intro}

Compared to today, computer vision was originally attempting to interpret images as projections of 3D scenes, not just processing 2D arrays of flat ``patterns''~\cite{do1961machine,minsky1969introduction,marr2010vision}.
In this pursuit, early research developed a series of intermediate tasks:
they focused on understanding optical properties like reflectance~\cite{wang1993layered,black1993framework}, 3D primitives through multi-view reasoning~\cite{marr1976cooperative,hartley2003multiple}, geometric reasoning through depth estimation~\cite{torralba2002depth}, instance recognition through visual correspondence~\cite{lowe1999object},  affordance through keypoint grounding~\cite{harris1988combined}, and forensics through intrinsic images~\cite{barrow1978recovering}.
% Yet in the modern era of large language models (LLMs), we, as a community, have focused less on such perceptual tasks, and have developed several new tasks \nascomment{this is vague; can you instead say ``and instead have developed new tasks, mostly expressed in natural language, emphasizing the vison-language connection learned by multimodal LLMs'' or something like that?} 
Yet in the modern era of large language models (LLMs), we, as a community, have focused less on such perceptual tasks, and instead have developed new tasks, mostly expressed in natural language, emphasizing the vision-language connection learned by 
multimodal LLMs~\cite{gpt4, team2023gemini, chen2023palix, liu2024visual, liu2024llavanext, alayrac2022flamingo, bai2023qwenvl, wang2023cogvlm, dai2023instructblip, chen2023sharegpt4v, internlmxcomposer2, lu2023unified}. 
%This is perhaps because many traditional computer vision tasks are not readily expressible using natural language, due to language's impreciseness 
% This might be because many traditional computer vision tasks cannot be easily described using natural language, due to the inherent imprecision of language (\eg, it is hard to precisely describe and locate keypoints within an image using language).
\xingyu{This might be because many traditional computer vision tasks resist mediation through natural language, due to the inherent imprecision of language (\eg, it is challenging to precisely pinpoint a spatial keypoint through language).}

This paper aims to highlight crucial aspects of visual perception that have been overlooked when evaluating multimodal LLMs.
To appropriately position our paper, let us revisit how we currently evaluate perception through using multimodal LLMs~\cite{liu2023mmbench, li2023seed, li2023seed2, yue2023mmmu, lu2023mathvista, liu2023visual, liu2023hidden}. 
While many of these benchmarks have been popularized as the de~facto evaluation measures for influential models like GPT-4V and Gemini-Pro, they conflate perception with language knowledge and reasoning.
At the risk of singling out one benchmark, let us consider two questions highlighted in the popular MMBench~\cite{liu2023mmbench}:
``\texttt{<image 1> Why is this hummingbird called ruby-throated?}'' and ``\texttt{<image 1> What will happen next? A: the person is gonna laugh B: the person is gonna cry.}'' 
For the first question, the vision subpart is to recognize the hummingbird. For the second, it only needs a coarse description of the image. Everything else is left to the language model to solve.
% While MMBech and other benchmarks have important contributions,
Such a conflation has also been reported for other benchmarks by previous work~\cite{yang2022empirical, hu2022promptcap, berrios2023towards}. Our experiments show that this conflation reductively evaluates perception as a dense captioning task. In other words, \textbf{by replacing the image with a task-agnostic dense caption, our experiments show that a ``blind'' GPT-4 performs well on these ``multimodal tasks''.}

\begin{figure}[t]
    \centering    
    \includegraphics[width=0.95\textwidth]{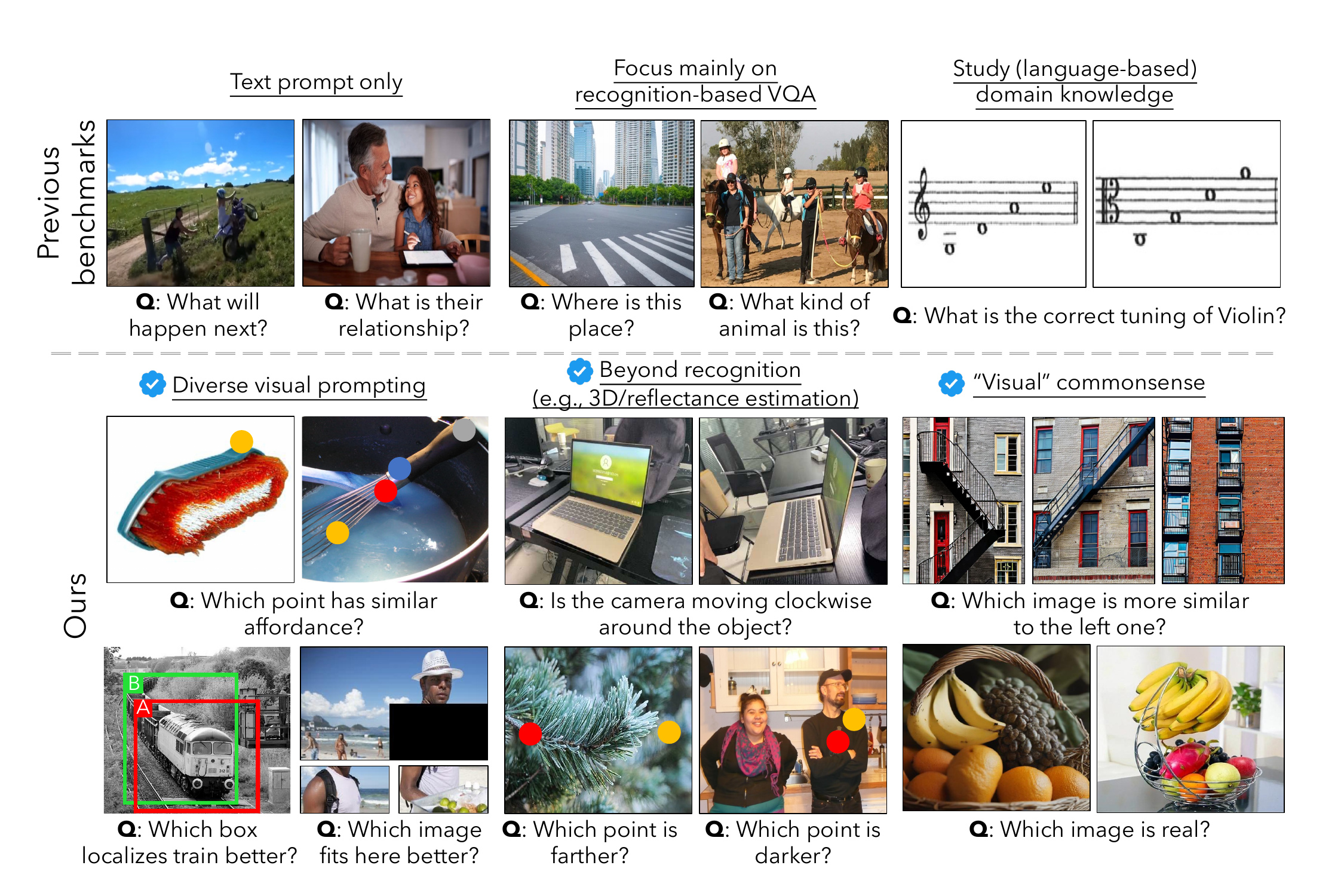}
    \caption{\textbf{Comparison between \bench and previous benchmarks.} \bench has several novel features: (1) \bench incorporates diverse visual prompts, like circles, boxes, and image masks, while previous benchmarks only have text questions and answers. (2) \bench evaluates a more comprehensive range of visual perception abilities, like multi-view reasoning, depth estimation, and reflectance estimation. Prior benchmarks are generally more focused on recognition-based VQA. (3) \bench contains ``visual'' commonsense problems that humans can answer within seconds, while prior benchmarks like~\cite{yue2023mmmu} require domain knowledge. The samples of previous benchmarks are from~\cite{liu2023mmbench, li2023seed, yue2023mmmu}. Part of our samples are curated from~\cite{lai2021functional, gupta2019lvis, fu-etal-2022-theres, ze2022category, chen2016single, bell14intrinsic, fu2023dreamsim}. 
    % \nascomment{top center:  ``where kind of animal is this'' is ungrammatical -- should it be ``what kind''?}
    }
    \label{fig:intro-difference}
    \vspace{-5mm}
\end{figure}

In response, we propose \bench. \bench reimagines traditional computer vision problems through a format that allows us to evaluate multimodal LLMs.
As partially demonstrated in~\Cref{fig:teaser},\footnote{\label{explainfootnote}The answers of the examples in \Cref{fig:teaser} are as follows. Relative depth: B; \jigsaw: A; \camerapose: right; \corr: A; \semCorr: C; \realness: final image; \iq: D; \dreamsim: upper one; \funcCorr: A; \iiw: they are about the same.} \bench consists of 14 classic computer vision tasks, ranging from low-level pattern matching (\eg, visual correspondences estimation) to mid-level reasoning (\eg, relative depth estimation), and extending to high-level visual understanding (\eg, visual similarity).
% It contains $3.9$K questions across $7.3$K images.
The image tasks are meticulously selected such that they are difficult to solve by reducing the evaluation using dense captioning; instead, the models must perceive the contents of the image(s) to answer.  
We recast each traditional task into a modern question-answering format, where answer choices are either images or text. 
\xingyu{ \bench contains $3.8$K questions across $7.3$K images, where questions may contain multiple images that are curated from a wide range of datasets~\cite{lin2014microsoft, krishna2017visual, bell14intrinsic, hpatches_2017_cvpr,chen2016single, gupta2019lvis, fu2023dreamsim, fu-etal-2022-theres}, encompassing indoor household scenes as well as outdoor urban or natural environments.
The questions and choices are either derived from the datasets, or manually written by humans. 
On average, each question can be solved by a human subject within a \bench of an eye, except the \iq. 
% \nascomment{``despite the IQ test'' sounds odd to me.  do you mean ``except the IQ test''?}
}
% Questions may contain multiple images, curated from a wide range of datasets~\cite{lin2014microsoft, krishna2017visual, bell14intrinsic, hpatches_2017_cvpr,chen2016single, gupta2019lvis, fu2023dreamsim, fu-etal-2022-theres}, encompassing indoor household scenes as well as outdoor urban or natural environments.
% The textual questions and choices are either derived from the datasets, or manually written by humans. On average, each question can be solved by a human subject within a \bench of an eye~.

We carefully evaluate 17 multimodal LLMs with various sizes (\ie, 7B, 13B, 34B) on \bench. We observe the paradox that \textbf{while these problems are easy for humans ({$95.70\%$} average accuracy), they are extremely hard for existing machinery} -- even GPT-4V model can only achieve $51.26\%$ accuracy on average, which is {$44.44\%$} worse than humans, 
and 13.17\% better than random guessing. 
% To validate whether scaling can mitigate the perception gap, we further compare open-sourced multimodal LLMs of various sizes, finding that their performance on \texttt{BLINK} is similar.
We also experiment with specialist vision models and find that they perform much better than multimodal LLMs. For example, the specialist outperforms GPT-4V by $62.8\%$ on visual correspondence estimation, $38.7\%$ on relative depth estimation, and $34.6\%$ on multi-view reasoning, in terms of absolute accuracy.
% Our experimental results suggest not only that we have been overestimating the perceptual capabilities of multimodal LLMs, but that perhaps they can learn from existing specialist models that already possess such capabilities.
% We believe \bench can serve as an effective testbed for bridging the gap traditional notions of perception with the modern generative capabilities of multimodal LLMs.
\xingyu{Our findings indicate that the perceptual abilities of multimodal LLMs have been previously overestimated. Furthermore, these models may benefit from integrating insights from specialized models that excel in these areas.
We believe \bench can serve as an effective testbed for bridging the gap between traditional notions of perception and the modern generative capabilities of multimodal LLMs.}

% We explore \bench and provide an comprehensive set of in-depth analysis, showing that: a) language proxies such as dense image captions are less useful on our benchmark versus others; b) different visual prompting can affect multimodal LLMs behaviors; c) specialist computer vision models can do much better than multimodal LLMs.
% We explore \bench and provide an comprehensive set of in-depth analysis, showing that: a) language proxies such as dense image captions are less useful on our benchmark versus others; b) different visual prompting can affect multimodal LLMs behaviors; c) specialist computer vision models can do much better than multimodal LLMs.
\section{Related Work}

\subsection{Multimodal Models}
Inspired by the impressive success in recent large language models (LLMs)~\cite{gpt3,gpt4, touvron2023llama, zheng2024judging, chowdhery2022palm}, a sequence of studies explore multimodal LMMs that can jointly understand vision and language information and generate textual answers through adding a modality adaption structure between a frozen visual encoder~\cite{radford2021learning,sun2023eva,fang2023eva} and a frozen LLM~\cite{touvron2023llama,zheng2024judging}.
Flamingo~\cite{alayrac2022flamingo} and BLIP-2~\cite{li2023blip} are two of the earliest works to explore these transformer-based multi-modality conjunction structures. They first pre-train on image-text matching datasets~\cite{lin2014microsoft,krishna2017visual,schuhmann2021laion,changpinyo2021cc12m} and then fine-tune on task-specific datasets such as visual question answer (VQA)~\cite{antol2015vqa,balanced_vqa_v2}.
Starting from LLaVA~\cite{liu2023improvedllava,liu2024llavanext}, people use LLM synthesized instruction-following chat data (which are in VQA format) for instruction tuning and achieve much better results~\cite{dai2023instructblip,chen2023minigptv2,team2023gemini,Qwen-VL}.
There have been extended studies that explore further capabilities of multimodal LLMs, especially on VQA reasoning~\cite{zellers2019vcr,hu2022promptcap,fu-etal-2023-generate,fu-etal-2022-theres,schwenk2022okvqa,hu2023tifa,fu2023interpretable,yan2024list}. However, they mainly focus on the textual reasoning abilities~\cite{wei2022chainofthought} within the multimodal LLMs and do not emphasize visual perceptions.

\subsection{Multimodal Benchmarks}

Traditional vision-language datasets are designed to assess single-task capabilities, such as optical character recognition (OCR)~\cite{liu2024hidden}, image captioning~\cite{lin2014microsoft}, and visual question answering~\cite{antol2015vqa,balanced_vqa_v2}. 
However, these datasets are often not comprehensive enough to holistically assess multimodal LMMs on general perception and reasoning abilities. 
Many recent papers have built more comprehensive benchmarks. 
MME~\cite{fu2023mme} is one of the earliest holistic benchmarks containing multi-modal Yes/No questions on the defined visual perception and language reasoning tasks. 
MM-Vet~\cite{yu2023mmvet} includes six sub-features from the previous datasets including recognition-focused questions, OCR, and math, providing a diverse while discrete evaluation set.
MMBench~\cite{liu2023mmbench} covers more subjects and provides a more robust circular evaluation setting. 
Seed-Bench~\cite{li2023seed,li2023seed2} benchmark has a more diverse source of inputs, including multiple-image inputs and video, and includes more tasks.
However, the visual perception questions in MME, MMBench, MM-Vet, and Seed-Bench are mainly extracted from existing VQA datasets or generated by GPT~\cite{gpt4} from image descriptions such as COCO-Caption~\cite{lin2014microsoft}, and are recognition focused, covering topics such as object (attribute)recognition, and OCR.
In contrast, we focus on multiple distinct nuanced perception abilities and recognition-level perception is only one of our focus.
Some other multimodal benchmarks have distinct focuses. MMMU~\cite{yue2023mmmu} aims at achieving expert-level artificial general intelligence by collecting domain-knowledge-required questions. 
HallusionBench~\cite{guan2023hallusionbench} mainly tests the language hallucination and visual
illusion phenomena.
MathVista~\cite{lu2023mathvista} presents exclusively mathematical domain visual questions based on images such as charts, tables, and diagrams. 
These benchmarks do not require human-level perception abilities as in \bench and therefore cannot measure model visual perceptions holistically.

% \vspace{-1.3em}
\section{The~\bench Benchmark}

Our goal is to faithfully evaluate the visual perception capabilities of existing Multimodal LLMs.
We seek to study the visual perception gap between humans and machineries, and offer deeper insights into potential pathways towards achieving more generalized machine perception.
% equip existing machinery with the ability to perceive our world as humans.
Based on the observation that existing benchmarks predominantly focus on evaluating visual recognition abilities, we introduce a novel benchmark,~\bench, designed to enable both quantitative and qualitative evaluation of the nuanced perception capabilities of multimodal LLM across various dimensions. 
We unfold this section by illustrating the overall design of \bench (\S\ref{sec:data_over}) and discussing its unique features comparing with previous benchmarks. 
\xingyu{Then we describe each task in detail, providing an in-depth explanation of the data curation process (\S\ref{sec:data_collect}). }
% Then we describe each task in detail and explain the data curation process (\S\ref{sec:data_collect}). 

\begin{table}[t]
\begin{minipage}{0.48\textwidth}
        \centering
        \includegraphics[width=\linewidth]{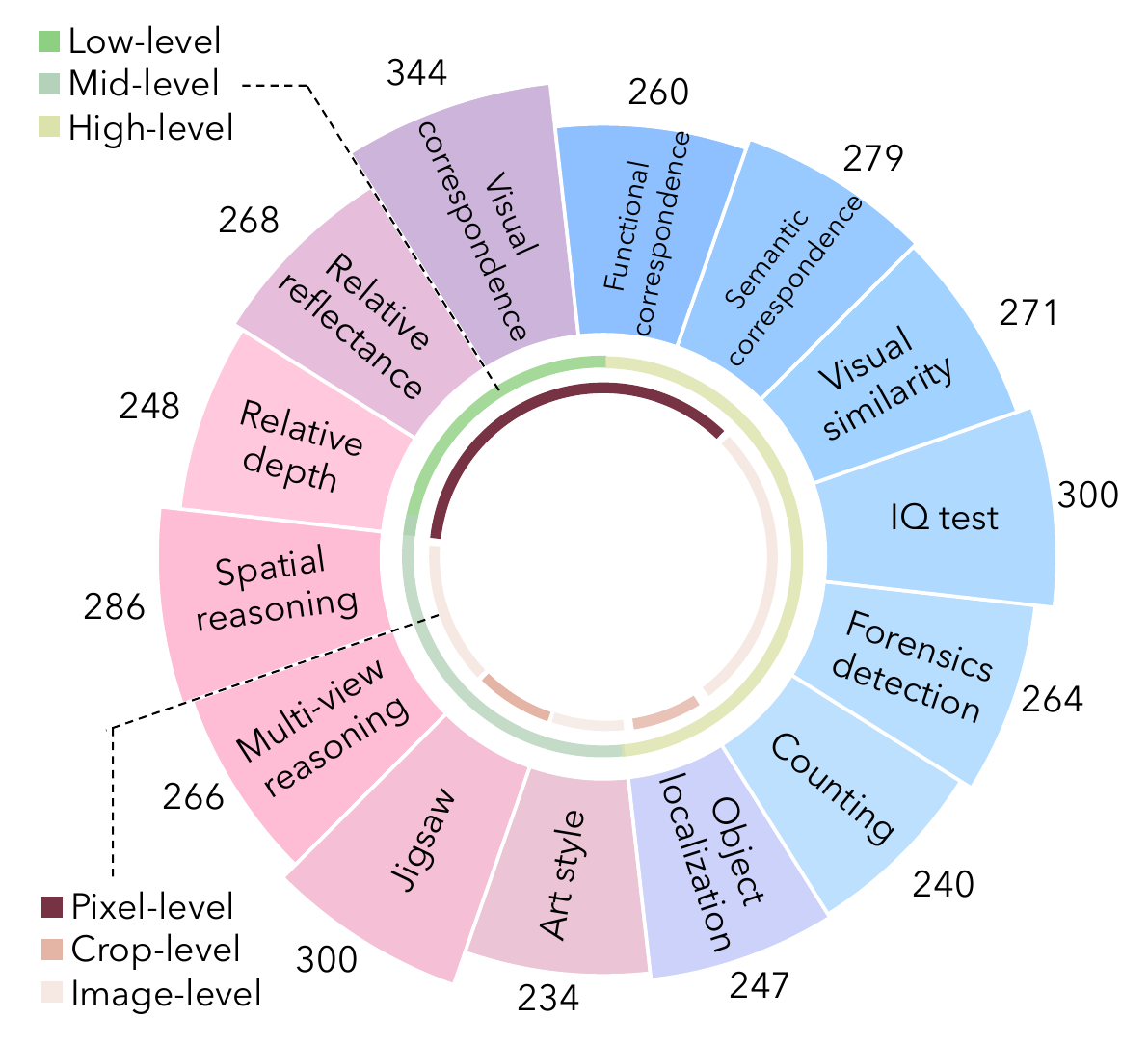}
        % \vspace{1mm}
        \captionof{figure}{Statistics of \bench. The benchmark includes 14 tasks, ranging from pixel-level to image-level perception, and low-level pattern matching (\eg, visual correspondences estimation) to mid-level reasoning (\eg, relative depth estimation), and extending to high-level visual understanding (\eg, visual similarity).}
        \label{fig:dataset-circular}
    \end{minipage}
    \hfill
    \begin{minipage}{0.48\textwidth}
        \centering
        \includegraphics[width=\linewidth]{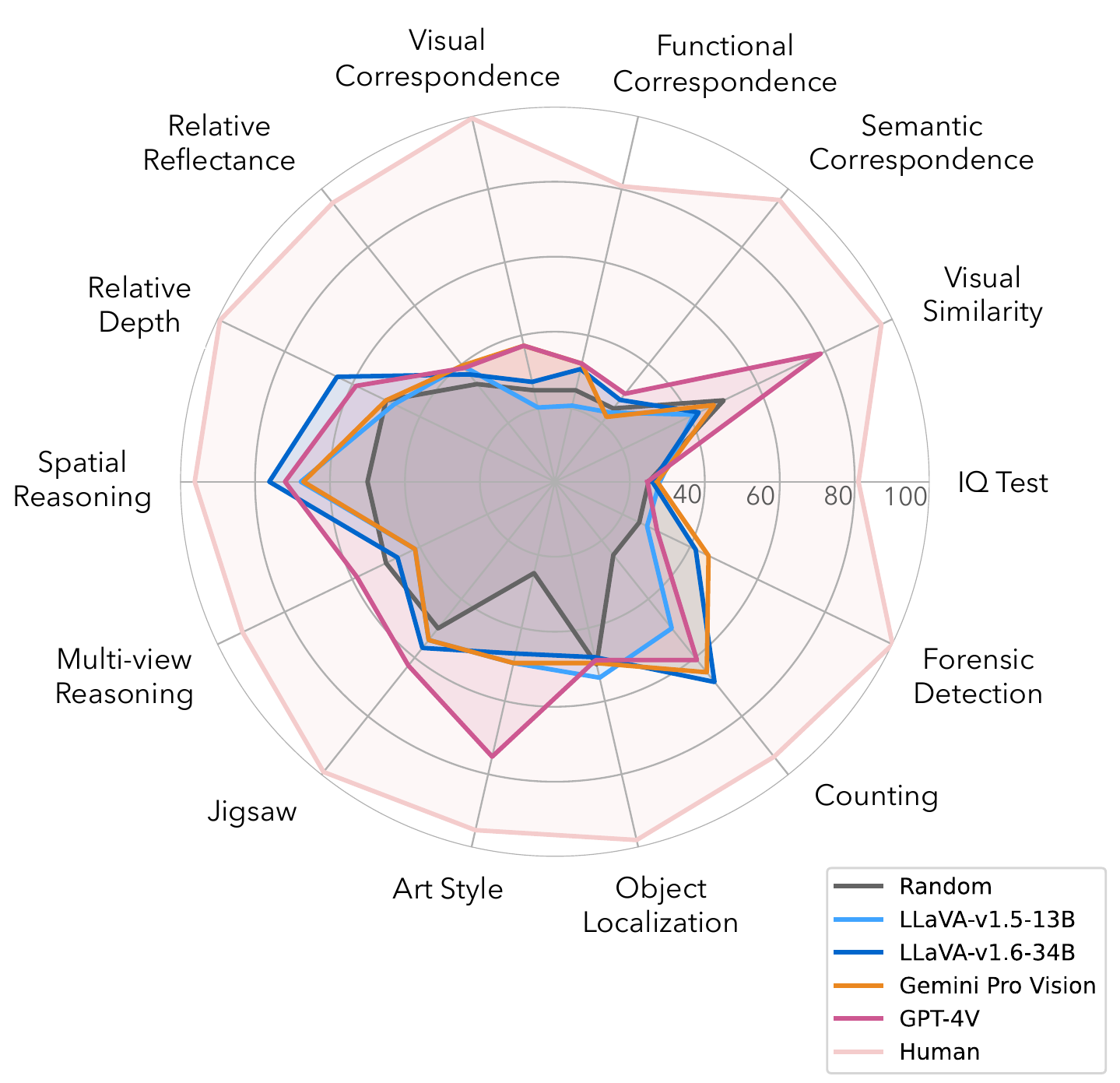}
        \captionof{figure}{Accuracies of multimodal LLMs on \bench test set. 
        % We also include random and human performance. 
        Please refer to \Cref{tab:main_results} and \S\ref{sec:exp_results} for more results and discussions.
        }
        \label{fig:radar}
    \end{minipage}
    \vspace{-5mm}
\end{table}

% In this section, we illustrate the overall design of \bench (\S\ref{sec:data_over}), then the detailed motivation and curation process of each task in the benchmark (\S\ref{sec:data_collect}).
% Our goal is to provide a novel benchmark to quantitatively and qualitatively evaluate the various levels of nuanced perception capabilities of multimodal LLMs. 

\subsection{Overview of \bench}
\label{sec:data_over}
To ensure that one can effectively measure what Multimodal LLMs can or cannot perceive, we carefully select 14 tasks  (see \S\ref{sec:data_collect} for the full list) that are difficult to solve by reducing the evaluation into text-only questions using dense captioning.
The tasks are drawn from either classic computer vision problems or recent applications of Multimodal LLMs, each of which requires a nuanced understanding of the visual data. 
They range from low-level pattern matching (\eg, visual correspondence) to mid-level spatial reasoning (\eg, relative depth), and up to high-level visual understanding (\eg, visual similarity). 
This variety allows for a systematic exploration of Multimodal LLMs' capabilities across different perceptual complexity layers.
Furthermore, these visual tasks vary in granularity, ranging from pixels (\eg, relative reflectance) to patches (\eg, jigsaw) and extending to the full image (\eg, forensic detection), enabling us to evaluate models' proficiency in observing at various scales.

To facilitate the evaluation of multimodal LLMs, we recast all tasks as multiple-choice question-answering problems. 
The options for answers may include images or texts, while the questions themselves can feature either single or multiple images. Prompts are designed to be both textual and visual in nature. 
We re-purposed several existing vision datasets as well as collected new data. In total, we contribute 3.9K multiple-choice questions and 7.3K images, with an even distribution between the validation and test sets. Numbers of each task are reported in~\Cref{fig:dataset-circular}, and more detailed statistics can be found in Appendix~\ref{app:data_stats}.

\vspace{1ex}\noindent\textbf{Key features of \bench: } 
Comparing with previous benchmarks, \bench has the following novel features: 
\begin{itemize}
    \item \textbf{Visual prompting}: 
    Unlike existing benchmarks that support only text prompting, \bench features a variety of visual prompts. This enables one to highlight specific areas within images, facilitating the evaluation of Multimodal LLMs' detailed understanding of these regions. 
    % Additionally, 
    It also offers an interface for researchers to investigate the impact of 
    % various 
    visual prompting techniques.

    \item \textbf{Perception beyond recognition}: Besides visual recognition, \bench considers a diverse set of visual perception abilities, such as 3D reasoning, geometric understanding, affordance reasoning, etc. The breadth allows one to evaluate Multimodal LLMs from an unique array of perspectives.

    \item \textbf{``Visual commonsense'' that does not require domain knowledge}:
    The questions in \bench are intentionally designed to be straightforward, requiring neither domain-specific knowledge nor expertise to answer. They are crafted in such a way that humans can solve them almost instantaneously, typically within a few seconds. This allows us to explore the fundamental gap in visual perception gap between humans and Multimodal LLMs, highlighting the paradox that problems easily solved by humans often pose significant challenges for machines.

    \item \textbf{Interleaved image-text formats}: \bench features a heterogeneous question-answering format, wherein both questions and choices can be presented as text or images. This diversity compels Multimodal LLMs to genuinely understand the questions, pushing the boundaries of their interpretative capabilities.

    \item \textbf{Diverse image sources}: \bench comprises a wide range of in-the-wild images sourced from various origins, covering everything from indoor and outdoor scenes to object-centric views and landscapes. This collection spans abstract diagrams, synthesized images, and authentic photographs, ensuring a comprehensive examination of visual perception

\end{itemize}
The design principles of \bench are also illustrated in \Cref{fig:intro-difference}. We will now describe each task in detail.

\subsection{Dataset Collection Process}

\label{sec:data_collect}

\bench comprises 14 tasks, all of which have been repurposed into a multiple-choice question-answering format. These tasks utilize a diverse collection of images from various sources, and we ensure that each test sample across all tasks features unique images.

\vspace{1ex}\noindent\textbf{Visual correspondence:} 
%Establishing correspondence between different images or within different parts of the same scene is crucial for tasks such as object tracking, 3D reconstruction, and image matching. 
% Visual correspondence estimation allows
This task aims to evaluate the ability of Multimodal LLMs to understand and identify the same scene point across various viewpoints, lighting conditions, or time. We exploit HPatches~\cite{hpatches_2017_cvpr} for this task. HPatches contains a number of image sequences, each of which are composed of images taken under different illuminations and/or viewpoints of a scene. 
For each question, we randomly sample two images and an interest point within them. Then we exploit the ground-truth homography to compute its correspondence. Finally, we randomly select three more interest points to serves as other choices.

%HPatches contains a number of image sequences of the same scene. Each sequence contains a reference image and 5 target images taken under a different illumination and/or viewpoint. We randomly sample one reference point on the reference image and use the provided ground-truth homography to compute its corresponding coordinates on a target image. We then sample three other points on the target image. The task is to choose which one of the four points corresponds to the reference point.

\vspace{1ex}\noindent\textbf{Relative reflectance: } This task aims to compare the reflectance (albedo) of two pixels. It allows us to evaluate Multimodal LLMs' understanding of material properties and their interaction with light, which is crucial for applications requiring high-fidelity visual interpretations. We curate our samples using human annotations from the Intrinsic Images in the Wild (IIW) dataset~\cite{bell14intrinsic}. Each question is based on an image and two specified points, with the objective being to identify which point is darker, or whether the two points have similar reflectance.

% Relative reflectance refers to the task of comparing the surface color of two points after being decomposed with shading and illumination effects. This task evaluates Multimodal LLMs' understanding of the material properties and their interaction with light, which is crucial for applications requiring high-fidelity visual interpretations. We curate our samples from the human annotations in the Intrinsic images in the Wild (IIW) dataset~\cite{bell14intrinsic}.

\vspace{1ex}\noindent\textbf{Relative depth: } Humans are good at judging relative depth~\cite{chen2016single}. This task can thus serve as a proxy to validate if the geometric understanding capabilities of existing multimodal LLMs are close to human. We curate our samples using human annotations from the Depth in the Wild~\cite{chen2016single} dataset. Each question contains an image and two specified points. The task is to determine which point is closer.

\begin{figure}[h!]
    \centering    \includegraphics[width=0.9\textwidth,trim={1.0cm 1.5cm 0cm 1.3cm},clip]{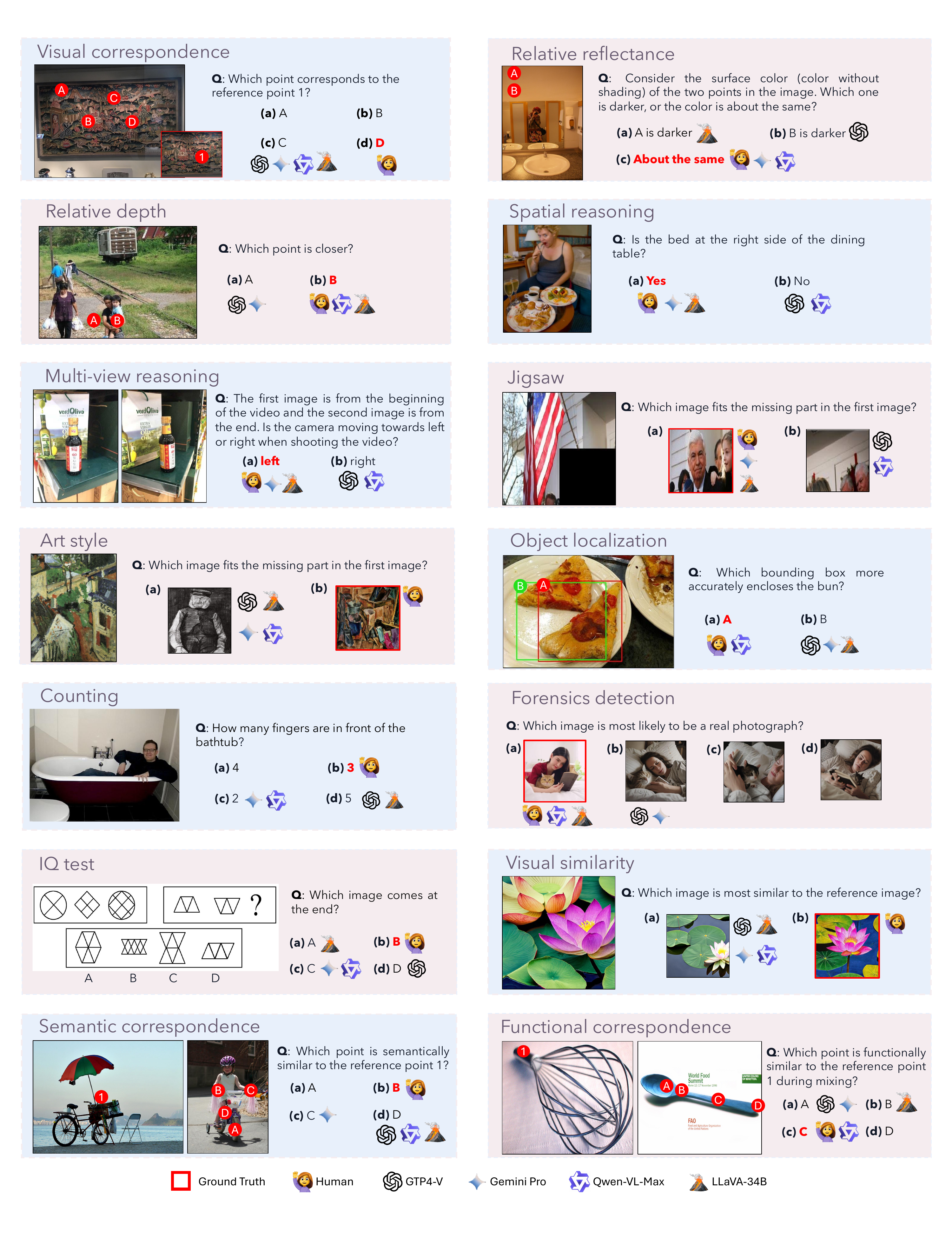}    
    \vspace{-2mm}
    \caption{ \textbf{Qualitative results} on \bench. For each task, we show the choice of LLaVA-v1.6-34B~\cite{liu2024llavanext}, Qwen-VL-Max~\cite{Qwen-VL}, Gemini Pro~\cite{team2023gemini}, GPT-4V~\cite{gpt4}, and humans. 
    Red choice indicates the ground truth. 
    Notice that the markers are intentionally enlarged for visualization purposes, and we make some images inset images to save space. 
    For IQ test, the third image is constructed by overlaying the first and second images.}
    \label{fig:qual}
    \vspace{-5mm}
\end{figure}

\vspace{1ex}\noindent\textbf{Spatial relation: } Understanding spatial relationships between objects in a scene is essential for interpreting complex visual environments. However, modern Multimodal LLMs often struggles with spatial concepts such as ``left'' and ``right''~\cite{yang2023dawn}. This task help us evaluate whether the models finally possess this vital skill. We curate our samples from the Visual Spatial Reasoning~\cite{liu2023visual} dataset. Each sample contains an image and a claim. The task is to determine if the claim is true or false. We reformat the claims into binary questions via GPT-3.5~\cite{gpt3}. 
%The Multimodal LLMs' task is to answer ``yes" or ``no".

\vspace{1ex}\noindent\textbf{Multi-view reasoning:} This task is centered on evaluating the multi-view reasoning capabilities of Multimodal LLMs. The objective is to deduce the relative camera motion based on two images of an object captured from different viewpoints. Our data is sourced from the Wild6D dataset~\cite{ze2022category}, which features videos of various objects recorded in diverse settings. We select two random frames from each video to calculate the relative camera motion. Recognizing that even humans might struggle to precisely articulate 3D motion details, we simplify the task by classifying motions into two broad categories: moving towards the left or moving towards the right. Despite the simplicity of these questions, as we will later demonstrate, they pose significant challenges for current models.

\vspace{1ex}\noindent\textbf{Jigsaw: } 
This task assesses the ability of Multimodal LLMs to recognize and group patterns, as well as to align patches based on continuity in shape, color, and texture. 
We utilize images from the TARA dataset \cite{fu-etal-2022-theres} and segment each of them into a 3x3 grid. We retain the three segments from the upper left corner as the reference image, and treat the central segment along with a randomly chosen segment as options. The objective is to identify the correct patch (\ie, the central patch).

\vspace{1ex}\noindent\textbf{Art style: } 
This task evaluates Multimodal LLMs capability to analyze and discern both local and global similarities in art styles among multiple images. Although there have been prior efforts to incorporate art-related questions into evaluation \cite{yue2023mmmu}, such attempts primarily focused on questions requiring expert-level knowledge, including deducing an artist's name and understanding historical contexts, rather than on direct image comparison. For this task, we collect paintings and their stylistic information from  \href{ https://www.wikiart.org/}{WikiArt}. Given one reference painting image and two other paintings as options, the model is tasked with identifying the one  that most closely shares the art style of the reference painting.

\vspace{1ex}\noindent\textbf{Object localization: } 
The ability to accurately detect and localize objects is critical for scene understanding. While previous benchmarks~\cite{liu2023mmbench} have explored this task, their focus was primarily on coarse localization. For instance, they might only ask the model if an object is located at the ``top'' or ``right'' side of an image.
\bench, in contrast, aims for a more fine-grained evaluation. We exploit images from LVIS~\cite{gupta2019lvis}, randomly sampling one object per image along with its ground-truth bounding box. Then we add Gaussian noise to the ground-truth box to create a confounding box. The goal is to select the correct one.

\vspace{1ex}\noindent\textbf{Counting: } 
% Counting is a well-known challenging problem for multimodal LLMs~\cite{yang2023dawn,hu2023visual}.
This task evaluate Multimodal LLMs' abilities in detection, recognition, and compositional reasoning, particularly in complex scenes where objects may overlap, be occluded, or vary in size and appearance. We select our questions from the TallyQA dataset \cite{acharya2019tallyqa}, known for its challenging human-written counting questions. Each sample comprises an image, a question, and a numerical answer. In addition to the correct answer, we randomly select three numbers to serve as confounding options.

\vspace{1ex}\noindent\textbf{Forensic detection: }
Recent advances in generative AI have raised concerns about malicious uses and have prompted calls for the automatic detection of fake content. To evaluate whether Multimodal LLMs can fulfill such a role, we construct sets of real and synthesized images that describe similar scenes and ask the models to identify the real ones. Specifically, we first generate synthetic images using Stable Diffusion XL~\cite{podell2023sdxl}, employing COCO captions~\cite{lin2014microsoft} as prompts. Then, we manually search online using these captions as descriptions and select high-quality photographs as the real images.

\vspace{1ex}\noindent\textbf{IQ test: }
This task evaluates the ability of Multimodal LLMs to engage in graphical reasoning, without requiring any domain-specific knowledge. We  manually collect test samples, along with human explanations, from various public, license-friendly online sources. Given visual examples and a selection of images, the objective is to identify the image that either continues the pattern established by the examples or is spatially consistent with them.

\vspace{1ex}\noindent\textbf{Visual similarity: }
This task aims to verify whether Multimodal LLMs possess a nuanced understanding of visual features, patterns, and aesthetics at a level comparable to humans. We select our samples from the DreamSim dataset \cite{fu2023dreamsim}. Given a reference image alongside two alternative images, the objective is to identify the image that most closely resembles the reference image in terms of visual similarity.

\vspace{1ex}\noindent\textbf{Semantic correspondence: }
This task focuses on identifying and matching semantically similar yet visually distinct elements across images, thereby evaluating the ability of Multimodal LLMs to understand the underlying semantics of object parts. Our samples are sourced from the SPair-71k dataset \cite{min2019spair}, which features pairs of images with multiple corresponding semantic points. For each task, we randomly select one semantic point in an image as a reference, and provide the matching point alongside three random semantic points in the paired image as options. The objective is to accurately identify the correct matches.

\vspace{1ex}\noindent\textbf{Functional correspondence: }
The task aims to identify points that are functionally similar across objects. 
It challenges Multimodal LLMs to extend their understanding beyond mere semantics, enabling them to infer the diverse functions an object can perform in various contexts. Such capability is crucial for applications in robotics. We derive our samples from the FunKPoint dataset \cite{lai2021functional}, which features paired images annotated for functional correspondences.
Following a method analogous to \semCorr, we present an action alongside two object images. One image includes a reference point, while the other offers four potential points. The objective is to select the point that best matches the reference in terms of functional affordances.

\vspace{1ex}\noindent\textbf{Data quality control: } 
To guarantee the quality of \bench, we manually go through all collected data and filter out data that are ambiguous.

\vspace{-3mm}
\section{Experiments}
\vspace{-3mm}

In this section, we first describe the experimental setup and the baselines (\S\ref{sec:exp_setup}).
Then we present a comprehensive evaluation of 16 recent multimodal LLMs (\S\ref{sec:exp_results}). We demonstrate that while humans can answer the questions with high accuracy, \bench is challenging for existing models.
%humans can answer its questions in the blink of an eye without domain-knowledge needed (\S\ref{sec:exp_results}). 
Finally, we provide detailed analyses on multiple experimental settings, including the effect of reducing images to captions, sensitivity to different visual prompts, and error analysis (\S\ref{sec:exp_analysis}).
% ensemble performance, the similarity between models, 
% Finally, we conduct various analyses on multiple experiment settings, including effect of reducing images to captions, visual prompting, ensemble, the similarity between models, along with error analysis(\S\ref{sec:exp_analysis}).

\vspace{-2mm}
\subsection{Experimental Setup}
\vspace{-2mm}
\label{sec:exp_setup}
\noindent\textbf{Multimodal LLMs: }
We evaluate \bench on 16 recent multimodal LLMs, including MiniGPT-4-v2~\cite{chen2023minigptv2}, OpenFlamingo-v2~\cite{awadalla2023openflamingo}, InstructBLIP (7B and 13B)~\cite{dai2023instructblip}, CogVLM~\cite{wang2023cogvlm}, LLaVA(v1, v1.5, v1.6, internLM, and xtuner versions, model size 7B, 13B, and 34B)~\cite{liu2024visual, liu2023improved, liu2024llavanext, internlmxcomposer2, 2023xtuner}, Yi-VL (6B and 34B)\footnote{More details are at the official website at~\url{https://www.01.ai/}\label{Yifootnote}}, Qwen-VL-MAX~\cite{Qwen-VL}, Gemini Pro~\cite{team2023gemini}, Claude 3 Opus~\cite{claude} and GPT-4V(vision)~\cite{gpt4}. 
%We refer the readers to \Cref{app:baselines} for more details. 
See \Cref{app:baselines} for more details.
% More details in \Cref{app:baselines}.

\vspace{1ex}\noindent\textbf{Evaluation setup: }
We follow standard setups as in the VLMEvalKit~\cite{2023opencompass}, where the temperature is set to 0 and retry is set to 10. However, we do not resize the images during any experiment.
For the models that do not support multiple images as input, we concatenate the images as input. 
We extract the choice from the models' output with a set of pre-defined rules and GPT-3.5-turbo~\cite{gpt3}. 
We refer the readers to Appendix \ref{appendix:details} for more details on visual prompting, how we generate the answers in \bench, and the human evaluation protocol. 
% In addition, we provide human evaluation scores and random guess scores for comparison\footnote{Note that the human score for \iq is annotated by authors does not reflect average human performance on this task.}.
% We follow standard setups as in the VLMEvalKit~\cite{2023opencompass}, where the temperature is set to 0 and retry is set to 10. However, we do not resize the images during any experiment.
% For the models that do not support multiple images as input, we concatenate the images as input, with more details in Appendix~\ref{app:concat_image}. 
% The prompts we use to generate the answers are provided in \bench and can be found in Appendix~\ref{app:prompt_task}, along with visual prompt settings.  
% We use GPT-3.5-turbo~\cite{gpt3} to extract answer choices from long answers, with details shown in Appendix \ref{app:prompt_eval}. We also provide rule-based choice extraction tools.
% In addition, we provide human evaluation scores and random guess scores for comparison. Details about human evaluation can be found in Appendix~\ref{app:human_eval}\footnote{Note that the human score for \iq is annotated by authors does not reflect average human performance on this task.}.

\begin{table}[h!]
    \centering
    % \small
    % \setlength{\tabcolsep}{1pt}
    \scalebox{0.85}{
{\fontsize{8.5pt}{10pt}\selectfont
    \begin{tabular}{lcccccccc} 
    \toprule[1.2pt]
    & \begin{tabular}{c} Validation \\$(1,901)$\end{tabular} 
    & \begin{tabular}{c} Test \\$(1,906)$\end{tabular} 
    & \begin{tabular}{c} \dreamsimshort \\$(136)$\end{tabular} 
    & \begin{tabular}{c} \countingshort \\$(120)$\end{tabular}
    & \begin{tabular}{c} \depthshort \\$(124)$\end{tabular} 
    & \begin{tabular}{c} \jigsawshort \\$(150)$\end{tabular} 
    & \begin{tabular}{c} \artshort \\$(117)$\end{tabular}
    & \begin{tabular}{c} \funcCorrshort \\$(130)$\end{tabular}    \\
    \midrule[1.2pt]
    Random Choice & 38.09& 38.09& 50& 25& 50& 50& 50 & 25\\
     Human & 95.67& 95.70& 96.70& 93.75 & 99.19 & 99.00& 95.30 & 80.77\\
    \hline \multicolumn{9}{c}{ \textbf{Open-source multimodal LLMs}} \\
    \hline 
    MiniGPT-4-v2~\cite{chen2023minigptv2} 
    & 34.23 & 34.57 & 52.94 & 10.83 & 49.19 & 26.00 & 47.86 & 18.46\\
    OpenFlamingo-v2~\cite{awadalla2023openflamingo} & 39.18 & 38.32 & 55.15 & 21.67 & 54.03 & 46.00 & 52.14 & 36.15\\
    InstructBLIP-7B~\cite{dai2023instructblip} & 39.72 & 38.65 & 46.32 & 29.17 & 50.81 & 54.00 & 47.86 & 23.85\\
    InstructBLIP-13B~\cite{dai2023instructblip} & 42.24 & 39.58 & 46.32 & 30.83 & 50.00 & 54.00 & 50.43 & 22.31\\
    LLaVA-internLM2-7B~\cite{2023internlm} & 37.71 & 36.06 & 52.94 & 52.50 & 52.42 & 34.67 & 30.77 & 23.08 \\
    Yi-VL-6B~\footref{Yifootnote} & 38.72 & 41.24 & 46.32 & 46.67 & 56.45 & 50.00 & 53.85 & 23.85 \\
    Yi-VL-34B~\footref{Yifootnote} & 41.68 & 42.78 & 50.00 & 58.33 & 53.23 & 54.00 & 46.15 & \textbf{39.23} \\
    LLaVA-v1.5-7B-xtuner~\cite{2023xtuner} & 39.36 & 40.81 & 46.32 & 53.33 & 50.81 & 54.00 & 47.86 & 23.85 \\
    LLaVA-v1.5-13B-xtuner~\cite{2023xtuner} & 42.00 & 41.31 & 46.32 & 45.00 & 54.03 & 53.33 & 47.86 & 26.15 \\
    CogVLM~\cite{wang2023cogvlm} & 41.54 & 39.38 & 46.32 & 38.33 & 50.81 & 52.67 & 49.57 & 23.85 \\
    LLaVA-v1.5-7B~\cite{liu2023improvedllava} & 37.13 & 38.01 & 46.32 & 43.33 & 51.61 & 11.33 & 47.86 & 21.54 \\
    LLaVA-v1.5-13B~\cite{liu2023improvedllava} & 42.66 & 40.55 & 46.32 & 50.00 & 47.58 & 54.00 & 47.86 & 20.77 \\
    LLaVA-v1.6-34B~\cite{liu2024llavanext} & 46.80 & 45.05 & 46.32 & \textbf{68.33} & 64.52 & 56.67 & 47.01 & 30.77 \\
    \hline \multicolumn{9}{c}{ \textbf{API-based models}} \\
    \hline Qwen-VL-Max~\cite{Qwen-VL} & 40.28 & 41.94 & 51.47 & 55.83 & 58.87 & 3.33 & 37.61 & 28.46 \\ 
    Gemini Pro~\cite{team2023gemini} & 45.16 & 45.72 & 55.88 & 65.00 & 50.00 & 54.00 & 49.57 & 32.31 \\
    Claude 3 OPUS~\cite{claude} & 44.05 & 44.11 & 70.59 & 49.17 & 57.26 & 32.67 & 60.68 & 22.31 \\
    GPT-4V(ision)~\cite{gpt4} & 51.14 & 51.26 & \textbf{83.09} & 60.83 & 58.87 & 62.67 & 78.63 & 31.54 \\
    GPT-4 Turbo~\cite{gpt4} & 54.61 & 53.89 & \textbf{83.09} & 60.83 & \textbf{66.94} & \textbf{66.00} & 81.20 & 31.54 \\
    GPT-4o~\cite{gpt4} & \textbf{60.04} & \textbf{59.03} & 65.44 & 51.67 & 64.52 & 58.00 & \textbf{82.91} & \textbf{39.23} \\
    \bottomrule
    \end{tabular}}}
    \vspace{2em}
    \setlength{\tabcolsep}{1.4pt}
    \scalebox{0.85}{
{\fontsize{8.5pt}{10pt}\selectfont
    \begin{tabular}{lcccccccc}
    \toprule[1.2pt]
    & \begin{tabular}{c} \semCorrshort \\$(140)$\end{tabular} 
    & \begin{tabular}{c} \spatialshort \\$(143)$\end{tabular}
    & \begin{tabular}{c} \localizationshort \\$(125)$\end{tabular} 
    & \begin{tabular}{c} \corrshort\\$(172)$\end{tabular} 
    & \begin{tabular}{c} \cameraposeshort \\$(133)$\end{tabular} 
    & \begin{tabular}{c} \iiwshort \\$(134)$\end{tabular} 
    & \begin{tabular}{c} \realnessshort\\$(132)$\end{tabular} 
    & \begin{tabular}{c} \iqshort\\$(150)$\end{tabular} \\
    \midrule[1.2pt]
    Random Choice & 25& 50& 50& 25& 50& 33.33&25 &25\\
     Human &96.07 & 98.25 & 98.00 & 99.42 & 92.48 & 95.14 & 100.00& 80.00\\
    \hline \multicolumn{9}{c}{ \textbf{Open-source multimodal LLMs}} \\
    \hline 
    MiniGPT-4-v2~\cite{chen2023minigptv2} & 26.43 & 51.75 & \textbf{56.00} & 23.84 & 52.63 & 31.34 & 17.42 & 19.33 \\
    OpenFlamingo-v2~\cite{awadalla2023openflamingo} & 23.57 & 46.85 & 52.00 & 25.00 & 41.35 & 43.28 & 15.91 & 23.33 \\
    InstructBLIP-7B~\cite{dai2023instructblip} & 25.00 & 55.24 & 44.80 & 22.67 & \textbf{58.65} & 29.85 & 29.55 & 23.33 \\
    InstructBLIP-13B~\cite{dai2023instructblip} & 22.86 & 64.34 & 52.00 & 20.93 & 54.14 & 46.27 & 13.64 & 26.00 \\
    LLaVA-internLM2-7B~\cite{2023internlm} & 22.14 & 74.13 & 48.00 & 21.51 & 41.35 & 32.84 & 3.79 & 14.67 \\
    Yi-VL-6B~\footref{Yifootnote} & 26.43 & 72.73 & 49.60 & 29.65 & 48.12 & 29.85 & 20.45 & 23.33 \\
    Yi-VL-34B~\footref{Yifootnote} & 21.43 & 70.63 & 54.40 & 23.84 & 41.35 & 46.27 & 17.42 & 22.67 \\
    LLaVA-v1.5-7B-xtuner~\cite{2023xtuner} & 24.29 & 74.83 & 45.60 & 23.84 & 42.11 & 26.87 & 36.36 & 21.33 \\
    LLaVA-v1.5-13B-xtuner~\cite{2023xtuner} & 22.14 & \textbf{77.62} & 48.00 & 22.09 & 41.35 & 46.27 & 29.55 & 18.67 \\
    CogVLM~\cite{wang2023cogvlm} & 23.57 & 67.13 & 43.20 & 20.93 & 57.14 & 26.87 & 24.24 & 26.67 \\
    LLaVA-v1.5-7B~\cite{liu2023improvedllava} & 32.14 & 70.63 & 48.80 & 20.35 & 49.62 & 36.57 & 28.03 & 24.00 \\
    LLaVA-v1.5-13B~\cite{liu2023improvedllava} & 23.57 & 67.83 & 47.20 & 20.35 & 41.35 & 45.52 & 27.27 & 28.00 \\
    LLaVA-v1.6-34B~\cite{liu2024llavanext} & 27.86 & 76.22 & 41.60 & 27.33 & 46.62 & 29.85 & 41.67 & 26.00 \\
    \hline \multicolumn{9}{c}{ \textbf{API-based models}} \\
    \hline Qwen-VL-Max~\cite{Qwen-VL} & 29.29 & \textbf{77.62} & 49.60 & 22.67 & 53.38 & \textbf{49.25} & 47.73 & 22.00 \\
    Gemini Pro~\cite{team2023gemini} & 22.14 & 67.13 & 46.40 & 37.21 & 41.35 & 46.27 & 45.45 & 27.33 \\
    Claude 3 OPUS~\cite{claude} & 20.71 & 57.34 & 46.40 & 31.40 & 57.89 & 27.61 & 62.12 & 21.33 \\
     % * concatenate images & 27.86 & - & - & 23.84 & 41.35 & - & 36.36 & - \\
    GPT-4V(ision)~\cite{gpt4} & 30.00 & 72.03 & 50.40 & 37.21 & 58.65 & 38.81 & 30.30 & 24.67 \\
    GPT-4 Turbo~\cite{gpt4} & 32.86 & 67.13 & 48.80 & 42.44 & 57.14 & 34.33 & 51.52 & \textbf{30.67} \\
    GPT-4o~\cite{gpt4} & \textbf{45.71} & 76.92 & \textbf{56.00} & \textbf{71.51} & \textbf{60.15} & 38.81 & \textbf{85.61} & \textbf{30.00} \\
    \bottomrule[1.2pt]
    \end{tabular}}}
    \vspace{-4mm}
    \caption{\textbf{Results of different models on the \bench test set}. The first row shows task names and number of test data. The best performance in each task is in-bold. For the sake of completion, we also show the average score on the validation set. Detailed scores on the validation set are in Appendix~\ref{app:analysis}.}
    \label{tab:main_results}
    \vspace{-10mm}
\end{table}

% \begin{figure}[h!]
%     \centering    \includegraphics[width=0.95\textwidth,trim={0cm 0 0.6cm 0},clip]{figures/qual.pdf}    
%     \vspace{-2mm}
%     \caption{ \textbf{Qualitative results} on \bench. For each task, we show the choice of LLaVA-v1.6-34B~\cite{liu2024llavanext}, Qwen-VL-Max~\cite{Qwen-VL}, Gemini Pro~\cite{team2023gemini}, GPT-4V~\cite{gpt4}, and humans. Notice that the markers are intentionally enlarged for visualization purposes. Also, we make some images inset images to save space. Red choice indicates the ground truth.}
%     \label{fig:qual}
%     \vspace{-5mm}
% \end{figure}

\vspace{-3mm}
\subsection{Main Results}
\label{sec:exp_results}

\noindent\textbf{Overall performance: } 
%\bench is extremely challenging to existing models. 
As shown in Table~\ref{tab:main_results}, the mean accuracy of 7B and 13B open-source Multimodal LLMs hover around 35--42\%, which is similar to random guess (38.09\%). The most proficient open-source model, LLaVA-v1.6-34B, achieves an accuracy of 45.05\%. Even the most advanced models, GPT-4V and Gemini Pro and Claude 3 OPUS, achieve accuracies of only 51.26\%, 45.72\%, and 44.11\% respectively. Their performance are merely 13.17\%, 7.63\% and 6.02\% better than random guessing and lag behind human performance by 44.44\%, 49.98\% and 51.59\%. Notably, for certain tasks such as \jigsaw,  \semCorr, \camerapose, \localization, and \iiw, some multimodal LLMs even underperform compared to random guessing. 
%We also observed that the performance variation of many baseline models are similar across tasks. We hypothesize it may be due to their reliance on the common encoder CLIP to process the visual information~\cite{radford2021learning}. 
%See Figure~\ref{fig:qual} for qualitative results.
Some qualitative results are shown in Figure~\ref{fig:qual}.

% The results in Table~\ref{tab:main_results} show that the mean performance of 7B and 13B open-source Multimodal LLMs is confined within a narrow range of 35-42\%, closely mirroring the outcomes expected from random guesses, which stand at 38.09\%. 
% LLaVA-v1.6-34B is the best-performing open-source model, achieving an accuracy of 46.40\%.
% Remarkably, even the leading-edge models, GPT-4V and Gemini Pro, only attain accuracies of 51.26\% and 45.72\% respectively on the test set. These figures represent modest improvements over random guessing—13.05\% and 7.63\% respectively—yet they fall significantly short of human benchmark performances by 43.65\% and 49.51\% on average. It is evident that the majority of tasks pose substantial challenges for the models, relegating their performance to levels barely above random guessing across tasks such as \depth, \jigsaw,  \semCorr, \localization, \camerapose, \iiw, and \realness.  
% We hypothesize that the uniformity in average performance across various tasks among many baseline models can largely be attributed to their reliance on the common visual encoder CLIP~\cite{radford2021learning}.
% We demonstrate the qualitative results on \bench as in Figure~\ref{fig:qual} with explanations included\footnote{The explanation for the IQ test in Figure~\ref{fig:qual} is that: The third figure is constructed by overlaying the first and second images.}.

\vspace{1ex}\noindent\textbf{In which tasks do multimodal LLMs show relative strengths and weaknesses?}
\Cref{fig:radar} shows the accuracies of the best-performing models on \bench: LLaVA-v1.6-34B~\cite{liu2024llavanext}, Gemini Pro~\cite{team2023gemini}, and GPT-4V~\cite{gpt4}. We observe that multimodal LLMs perform relatively better on spatial reasoning, art style, and counting tasks, in which they are much better than random guessing. The models also demonstrate some capability in relative depth and forensics detection. Overall, they are doing relatively well on mid-level perception tasks. In terms of granularity, the models in general perform better on image-level tasks and struggle on pixel-level and crop-level tasks.

\vspace{1ex}\noindent\textbf{GPT-4V behaves differently: }
\Cref{fig:radar} and \Cref{tab:main_results} show an interesting phenomenon: GPT-4V's performance pattern is different from other models. Compared with its counterparts, GPT-4V is much better in visual similarity, art style, jigsaw, and multi-view reasoning. Specifically, its performance on visual similarity is 29\% better than Gemini Pro, demonstrating that GPT-4V possesses a nuanced understanding of visual patterns and aesthetics that is similar to humans. In contrast, Gemini Pro and LLaVA have similar performance patterns. 
%We hypothesize that these models may have undergone similar training processes.

% \vspace{1ex}\noindent\textbf{How to deal with multiple-image inputs?}
% Among all baseline models, only GPT-4V and Gemini Pro accept multi-image inputs. 
% There are two ways to process these inputs, either directly sending multiple images, or first concatenating multiple images into a single composite image and then sending the single image.
% We experiment with both options to see their impacts on the benchmark performance. The results are in Table~\ref{tab:main_results}. 
% We find that GPT-4V shows a consistent decline in performance across all tasks when taking concatenated images as input. However, the impact of concatenating images to Gemini Pro is task-dependent, with the performance decreasing in most tasks while increasing in \art and \semCorr. We believe this variation in model response could potentially indicate the difference between the design natures of the model structures and training.

\vspace{1ex}\noindent\textbf{Human performance:}
Human evaluators achieve over 95\% accuracy across most tasks, with an average accuracy of 95.70\% .\footnote{Note that the human score for \iq is annotated by authors. It may not reflect typical human performance, which is also expected to vary.} 
This performance disparity between humans and multimodal LLMs highlights the significant visual perception gap that exists between current machine learning models and humans in perceiving, processing, and understanding complex visual and textual context.

% Humans achieve over 90\% performance on almost all of these tasks, and achieves 94.97\% accuracy in average\footnote{Note that the human score for \iq is annotated by authors and does not reflect average human performance on this task.}.
% Conversely, human evaluators consistently achieve superior performance, exceeding 95\% across most of the tasks and achieving an average accuracy of 95.70\%. This disparity underscores the significant visual perception gap that exists between current machine learning models and humans in perceiving, processing, and understanding complex visual and textual context.

\subsection{Analysis}
\label{sec:exp_analysis}
% \vspace{-5mm}

\begin{table}[t]
\begin{minipage}{0.4\textwidth}
        \centering
        \includegraphics[width=\linewidth]{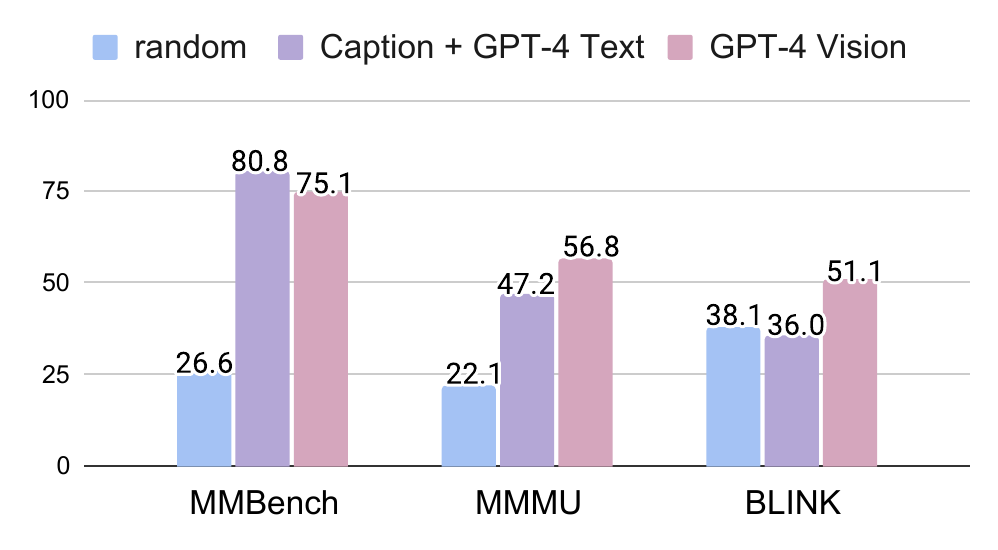}
        \captionof{figure}{Performance of using image caption + text-only GPT-4 \vs GPT-4 Vision on MMBench~\cite{liu2023mmbench}, MMMU~\cite{yue2023mmmu}, and \bench (\S\ref{sec:exp_analysis}).
        }
        \label{fig:caption}
        \vspace{-2mm}
    \end{minipage}
    \hfill
    \begin{minipage}{0.56\textwidth}
        \centering
        \vspace{2mm}
        \includegraphics[width=\linewidth]{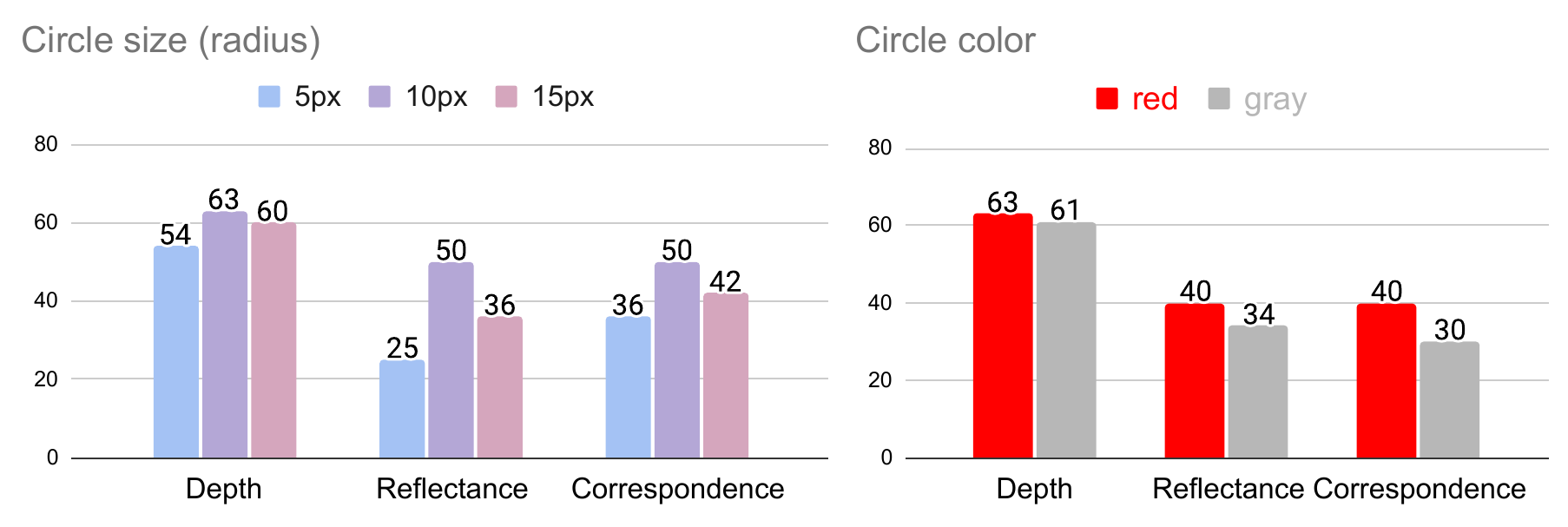}
        % \vspace{0.1mm}
        % \captionof{figure}{The accuracy of GPT-4V when using visual prompts with different circle sizes and colors on \depth, \iiw, and \corr tasks in \bench. 
        \captionof{figure}{Accuracy of GPT-4V with different visual prompts (\eg, different circle sizes, colors) on \depth, \iiw, and \corr tasks. More discussions in \S\ref{sec:exp_analysis}. 
        }
        \label{fig:visual_prompting}
        \vspace{-2mm}
    \end{minipage}
    \vspace{-5mm}
\end{table}
\noindent\textbf{Is dense captioning all you need for a multimodal LLM benchmark?}
To answer the question, we reduce multimodal benchmarks to a text-only problem. Specifically, we convert images into task-agnostic dense image captions with GPT-4V. The dense caption describes detailed information about the image and the visual prompts (\eg, where each circle is), using language. For each multimodal question, we prompt the text-only GPT-4-0125-preview model with image captions and the textual question and evaluate if the ``blind'' GPT-4 can answer the question. We call this \texttt{Caption + LLM}.
This experiment is predicated on the hypothesis that captioning involves predominantly recognition-centric perception.
If using captions along with text-only LLMs yields performance comparable to or surpassing that achieved through the integration of images with multimodal LLMs, 
then the perception demands of that benchmark are primarily confined to recognition only.

We experiment with \bench, MMBench~\cite{liu2023mmbench} and MMMU~\cite{yue2023mmmu}, as illustrated in Figure~\ref{fig:caption}. 
Surprisingly, we find that the \texttt{Caption + LLM} setting achieves better results on MMBench than GPT-4V (with 5.7\% increase in accuracy). On MMMU, \texttt{Caption + LLM} achieves 47.2\% accuracy, which is 9.6\% lower than GPT-4V performance, but is still much better than random guessing. On \bench, \texttt{Caption + LLM} fails, achieving random guessing performance.
These results indicate that dense captions cover the visual information needed for MMBench. For MMMU, image captions carry a large portion of visual information needed to answer the domain-knowledge-specific questions. Meanwhile, the performance decrease observed in \bench suggests the necessity for advanced perceptual abilities beyond what is currently attainable with general captions. This variance highlights the limitations of existing multimodal LLM benchmarks in addressing the full spectrum of visual perception.

\vspace{1ex}\noindent\textbf{Effect of visual prompting on \bench: }
Several \bench tasks involve visual prompting. Prior work~\cite{shtedritski2023does} shows that factors like shape, size, and color may affect task performance, and circles give the best overall performance. Following \cite{shtedritski2023does}, we adopt circles in \bench and analyze the effect of circle sizes and colors on multiple tasks in \Cref{fig:visual_prompting}. We experiment with relative depth, relative reflectance, and visual correspondence, with 100 validation set samples per task. The images are all reshaped to 1024px height. We experiment with circles with 5px, 10px, and 15px radius, and with red or gray color. We find that red is better than gray for all tasks. Also, the optimal circle size is task-dependent. On average 10px circles work the best, and we use it for all evaluations in this paper. The experiments suggest that visual prompting can have a big impact on multimodal LLM performance, and improving visual prompts or improving model robustness to prompt variation is a promising direction for future research~\cite{yang2023setofmark}.

\begin{table}[t]
\small
\centering
% \vspace{-1em}
\scalebox{0.85}{
{\fontsize{8.5pt}{10pt}\selectfont
\begin{tabular}{l|cccccc}
\toprule[1.2pt]
% & \multicolumn{7}{|c|}{Instruction-tuning tasks} & \multicolumn{3}{c}{Zero-shot tasks} \\
% \midrule
Task & \corrshort & \depthshort  & \cameraposeshort & \semCorrshort & \realnessshort & \iiwshort\\
\midrule
Random & 25.00 &  50.00 & 50.00 & 25.00 & 25.00 & 33.33 \\
Human & 99.56 & 99.59 & 92.10 & 94.60  & 100.00 & 99.63\\
\hline
Gemini Pro & 42.44 & 40.32 & 44.36 & 26.62 & 50.76& 45.52\\
GPT-4V & 33.72 & 59.68 & 55.64 & 28.78 & 34.09 & 38.81\\
\hline
Specialist & DIFT~\cite{tang2023emergent} & DepthAnything~\cite{depthanything}  & LoFTR~\cite{sun2021loftr} & DIFT~\cite{tang2023emergent} & DIRE~\cite{wang2023dire} & Ordinal Shading~\cite{careagaIntrinsic}\\
& 96.51 & 97.58 &  90.22 & 71.22 & 68.94 & 77.61\\
\bottomrule[1.2pt]
\end{tabular}
}
}
\vspace{1.5mm}
\caption{
Comparison between multimodal LLMs, specialists, and human performance on the \bench dev set.
The specialists perform much better than multimodal LLMs.
}
\vspace{-5mm}
\label{tab:specialist}
% \bottomrule
\end{table}

\vspace{1ex}\noindent\textbf{Can specialist models solve \bench tasks?}
Specialists can serve as a proxy upper bound of how good multimodal LLMs could be. 
We download the trained checkpoints for six specialist models and evaluate them on \bench.
As shown in \Cref{tab:specialist}, the specialists perform much better than GPT-4V and Gemini Pro, outperforming the best multimodal LLM by 18\% to 57\% on these tasks. Specifically, DepthAnything~\cite{depthanything} and DIFT~\cite{tang2023emergent} achieve human-level performance on depth estimation and visual correspondence, whereas multimodal LLMs fail miserably. 
This sheds light on the possibility that multimodal LLMs may progress on these tasks given the correct data and training strategy. For instance, one possible way is to distill existing specialist models into multimodal LLMs~\cite{hu2023visual}.

% Given that existing multimodal LLMs perform poorly on \bench tasks, we would like to explore ways to improve them. The first question to answer is whether it is possible to train deep learning models that excel on these tasks with existing data. To address this question, we explore if existing specialist CV models could solve tasks in \bench.
% \Cref{tab:specialist} shows the performance of specialist models on six tasks and compares them with GPT-4V, Gemini Pro, and human performance. We directly downloaded the trained checkpoint from each work and test on \bench. The specialists perform much better than GPT-4V and Gemini, outperforming the best multimodal LLM by 18\% to 57\% on these tasks. Specifically, DepthAnything~\cite{depthanything} and DIFT~\cite{tang2023emergent} achieve close-to-human performance on depth estimation and visual correspondence, while multimodal LLMs fail miserably. This sheds light on the possibility that multimodal LLMs may progress on these tasks given the correct data and training strategy. One  way is to distill existing specialist models into multimodal LLMs, as proposed in~\cite{hu2023visual}.

\vspace{1ex}\noindent\textbf{Error analysis of GPT-4V:} 
We randomly sampled 140 error instances made by GPT-4V on \bench, 10 per task, and meticulously examined them. 
% We provide a summary of errors here with more details in \Cref{app:error_analysis}.
% Overall, we find that GPT-4V has limited capability in understanding fine-grained visual details, reasoning about multiple images, and localization. 
% Although the model sometimes rejects to answer (6.4\% of errors) when it is unsure, GPT-4V is still prone to hallucination when conducting cross-modal reasoning.
% The most common types of hallucinations are listed as follows:
The most common types of errors are:
\noindent\textbf{Hallucinate fine-grained patterns and attributes} (24.2\%): the model hallucinates the nuanced details of objects. This error is most common for \iiw, \realness, and \jigsaw tasks.
\noindent\textbf{Hallucinate visual prompt locations} (20.0\%): the circle location described by the model is wrong. This is common for \corr and \depth tasks.
% \noindent\textbf{Spatial relation errors (14.3\%)}: GPT-4V fails to identify the correct spatial relations between the objects.
% \noindent\textbf{Reasoning errors/Hallucination (12.9\%)}: while the model correctly interprets the image and the question, it fails to derive the correct answer.
% \noindent\textbf{Other hallucinations}: failures on capturing overall setting/style (8.6\%) and errors on grounding an object (5.7\%). 
Other errors include Failures on capturing overall setting or style (8.6\%), and Failures on grounding an object (5.7\%). 
More details are in \Cref{app:error_analysis}.
\vspace{-3mm}
\section{Conclusion}
\vspace{-3mm}

We introduced \bench, a new multimodal LLM benchmark that evaluates core visual perception abilities not found in existing evaluations. While these tasks seem trivial for humans to solve “within a blink”, we find they pose significant challenges for current multimodal LLMs. Even the powerful GPT-4V and Gemini models only achieve around 50\% accuracy on Blink, far below the 95.7\% human performance. 
We conduct extensive analysis, measuring the effect of converting images to dense captions, visual prompting, self-consistency, analyzing the capabilities of specialist models, and conducting error analysis.
We highlight that specialist computer vision models are performing much better than GPT-4V and Gemini on \bench,  shedding light on the possibility that multimodal LLMs may have big progress on these tasks. 
Ultimately, Blink provides a simple yet effective testbed for multimodal LLMs to catch up with human-level visual perception.

\bibliographystyle{splncs04}
\bibliography{references}
\clearpage

\appendix

\noindent In the supplemental materials,~\Cref{appendix:details} contains additional details on \bench dataset collection
and model inference,~\Cref{app:baselines} provides more details of the baseline models,~\Cref{app:analysis} includes experimental analyses on \bench, and~\Cref{app:limit} discusses limitations.
% Additional visualizations of \bench examples are available on our project page.

\section{\bench Details}
\label{appendix:details}

% We show qualitative examples of actual \bench prompts, images, visual prompts, and GPT-4V answers in \Cref{fig:real_examples1,fig:real_examples2,fig:real_examples3,fig:real_examples4,fig:real_examples5,fig:real_examples6,fig:real_examples7}.

\subsection{Visual Prompts Details}
\label{app:visual_prompt}
\begin{figure}[h]
    \centering
    \includegraphics[width=0.95\textwidth,trim={0 8cm 0 0},page=1]{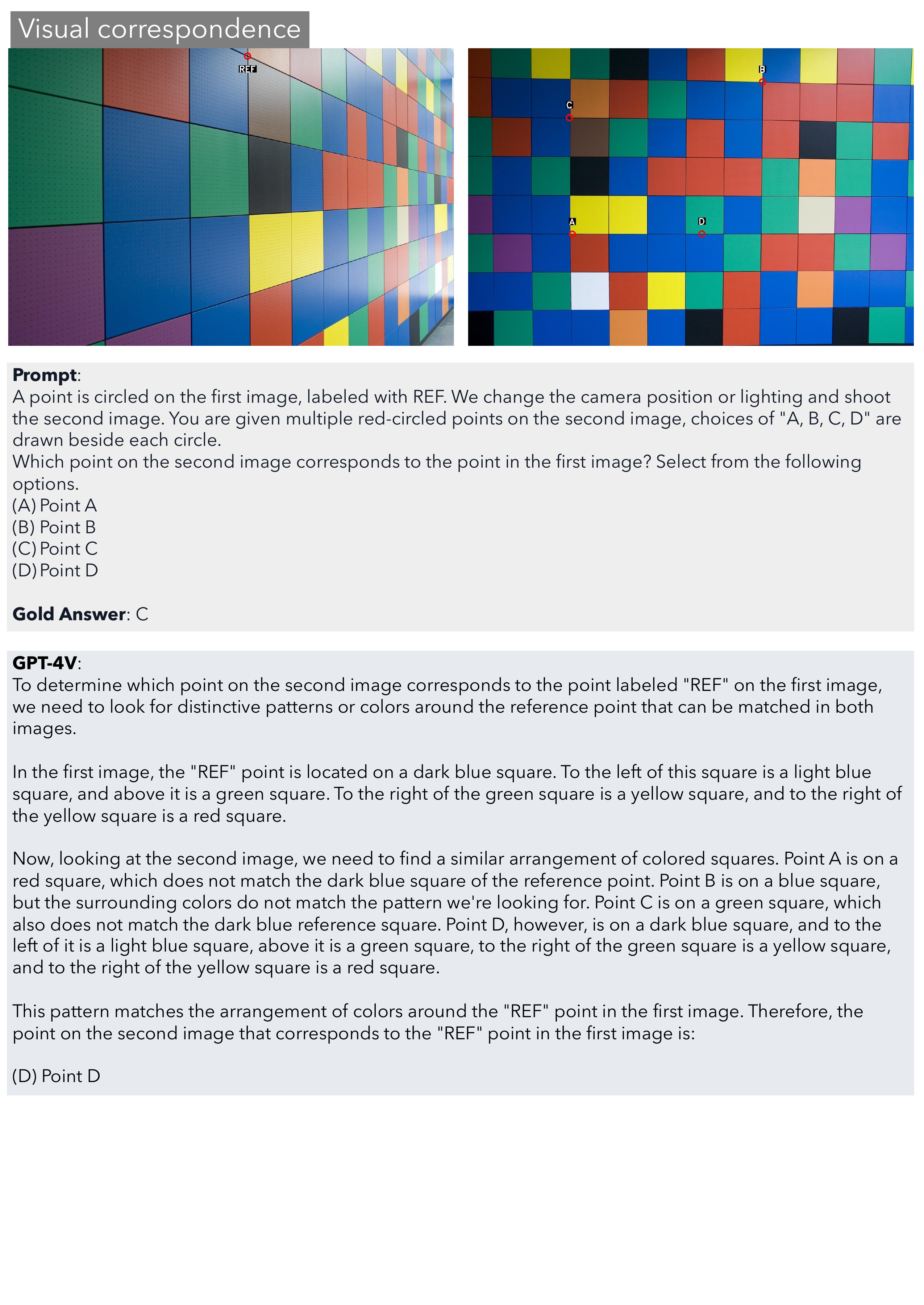}
    \vspace{-2mm}
    \caption{Examples of actual-sized data in \bench with GPT-4V predictions.(1/11)
    }
    \label{fig:real_examples1}
\end{figure}

There are three types of visual prompts in \bench: circles, boxes, and masks as shown in \Cref{fig:qual}. 
As for \corr, \funcCorr, \semCorr, the red circles have radius 10px on images resized to 1024px height.
For \iiw, we draw white circles to avoid color confusions.
For \localization, the boxes are in red and green.
For \jigsaw, the masks are kept black.
Since the examples in \Cref{fig:qual} are different from the actual ones for illustrative purposes, we show some actual-sized example data as in \Cref{fig:real_examples1,fig:real_examples2,fig:real_examples3,fig:real_examples4,fig:real_examples5,fig:real_examples6,fig:real_examples7,fig:real_examples8,fig:real_examples9,fig:real_examples10,fig:real_examples11}, with GPT-4V predictions attached.

\begin{figure}[h]
    \centering    
    \includegraphics[width=0.95\textwidth]{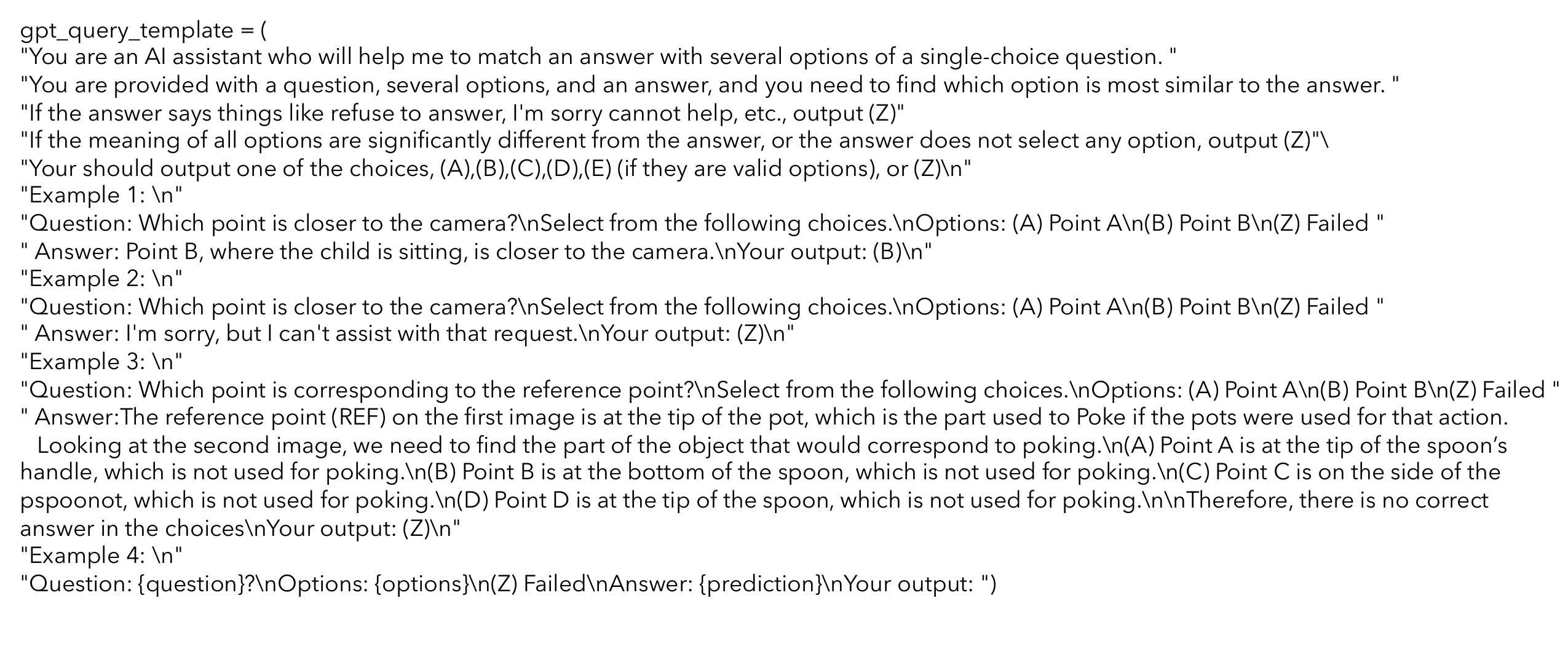}
    \vspace{-2mm}
    \caption{The evaluation prompts used for option label extraction.}
    \label{fig:eval_prompts}
\end{figure}

% \includepdf[width=0.95\textwidth,pages=-]{figures/example.pdf}

\begin{figure}[h]
    \centering
    \includegraphics[width=0.95\textwidth,trim={0 0 0 0},page=2]{figures/example.pdf}
    % \vspace{-2mm}
    \caption{Examples of actual-sized data in \bench with GPT-4V predictions. (2/11)
    }
    \label{fig:real_examples2}
\end{figure}

\begin{figure}[h]
    \centering
    \includegraphics[width=0.95\textwidth,trim={0 0 0 0},page=3]{figures/example.pdf}
    \vspace{-2mm}
    \caption{Examples of actual-sized data in \bench with GPT-4V predictions. (3/11)
    }   
    \label{fig:real_examples3}
\end{figure}

\begin{figure}[h]
    \centering
    \includegraphics[width=0.95\textwidth,trim={0 24cm 0 0},page=4]{figures/example.pdf}
    \vspace{-2mm}
    \caption{Examples of actual-sized data in \bench with GPT-4V predictions. (4/11)
    }
    \label{fig:real_examples4}
\end{figure}    

\begin{figure}[h]   
    \centering
    \includegraphics[width=0.95\textwidth,trim={0 16cm 0 0},page=5]{figures/example.pdf}
    \vspace{-2mm}
    \caption{Examples of actual-sized data in \bench with GPT-4V predictions. (5/11)
    }
    \label{fig:real_examples5}
\end{figure}  

\begin{figure}[h]
    \centering
    \includegraphics[width=0.95\textwidth,trim={0 0 0 0},page=6]{figures/example.pdf}
    \vspace{-2mm}
    \caption{Examples of actual-sized data in \bench with GPT-4V predictions. (6/11)
    }
    \label{fig:real_examples6}
\end{figure}

\begin{figure}[h]
    \centering
    \includegraphics[width=0.95\textwidth,trim={0 0 0 0},page=7]{figures/example.pdf}
    \vspace{-2mm}
    \caption{Examples of actual-sized data in \bench with GPT-4V predictions. (7/11)
    }
    \label{fig:real_examples7}
\end{figure}

\begin{figure}[h]
    \centering
    \includegraphics[width=0.95\textwidth,trim={0 0 0 0},page=8]{figures/example.pdf}
    % \vspace{-2mm}
    \caption{Examples of actual-sized data in \bench with GPT-4V predictions. (8/11)
    }
    \label{fig:real_examples8}
\end{figure}

\begin{figure}[h]
    \centering
    \includegraphics[width=0.95\textwidth,trim={0 18cm 0 0},page=9]{figures/example.pdf}
    \vspace{-2mm}
    \caption{Examples of actual-sized data in \bench with GPT-4V predictions. (9/11)
    }
    \label{fig:real_examples9}
\end{figure}

\begin{figure}[h]
    \centering
    \includegraphics[width=0.95\textwidth,trim={0 19cm 0 0},page=10]{figures/example.pdf}
    \vspace{-2mm}
    \caption{Examples of actual-sized data in \bench with GPT-4V predictions. (10/11)
    }
    \label{fig:real_examples10}
\end{figure}

\begin{figure}[h]
    \centering
    \includegraphics[width=0.95\textwidth,trim={0 0 0 0},page=11]{figures/example.pdf}
    \vspace{-4mm}
    \caption{Examples of actual-sized data in \bench with GPT-4V predictions. (11/11)
    }
    \label{fig:real_examples11}
\end{figure}

% \paragraph{\iq}
% IQ test is not a classical computer vision problem, but there has been some related studies on intelligence test with vision models~\cite{benny2021scale}, focusing on problems that do not require prior knowledge in language, reading, or arithmetics.
% This task tests multimodal LLMs' ability to solve graphical reasoning problems that do not require any domain-knowledge.
% We manually collect test samples along with human explanations from multiple public online sources such as \href{https://wenku.baidu.com/view/1456a3165b0102020740be1e650e52ea5518ce9b.html?fr=income4-doc-search&_wkts_=1709522468939&wkQuery=%E4%B8%AD%E5%9B%BD%E5%85%AC%E5%8A%A1%E5%91%98%E5%9B%BD%E8%80%83%E5%9B%BE%E5%BD%A2%E6%8E%A8%E7%90%86%E9%A2%98+100%E9%81%93&needWelcomeRecommand=1}{Baidu Wenku} using translation.
% We include two types of problems in this task, with ratio being 9:1: 1) Pattern following: identify the image that follows the same pattern or rule established by the previous set of image. 2) Spatial imagination: determine which image can be folded from the outer surface of the displayed carton (see Fig. \ref{fig:tasks2}).
% % In the val/test set, 90\%/91.3\% of the data is for pattern recognition and 10\%/8.7\% is for spatial recognition. 
% The task is given visual examples and choices of images, select the one that follows the pattern from examples or is spatially consistent with the examples.

% \subsection{Prompts used for each task}
% \label{app:prompt_task}

\subsection{Spatial Relation Curation Process}
We curate our samples from the Visual Spatial Reasoning~\cite{liu2023visual} dataset. Each original sample contains an image and a claim, which is either true or false. One example being ``\texttt{Caption: The cow is ahead of the person. Label: False.}''
We reformat the claims into binary questions via GPT-3.5~\cite{gpt3}, \eg ``\texttt{Question: Is the cow ahead of the person? Choices: (A) Yes (B) No  Label: (B)}''

\subsection{Evaluation Prompts}
\label{app:prompt_eval}

Following MMBench~\cite{liu2023mmbench}, given model outputs, we first try to extract choices with exact matching (\eg, for ‘C’, we try to match "C" and "(C)", etc). 
If failed, we extract the choices using GPT-3.5~\cite{gpt3}. We provide GPT with the question, options, and model prediction, and then request GPT to align the prediction with one of the given options, or ``Z'', meaning that it fails to match to an option. Screenshot of the prompts we used are in~\Cref{fig:eval_prompts}.

\subsection{Human Evaluation Protocol}
\label{app:human_eval}
We assign two humans (coauthors) for each task in~\bench and present their average scores as human performance. The human agreement scores range between 80-99\%, with the lowest being on \art and \funcCorr, highest being \depth, \localization, and \realness.
Notice that the only exception is the \iq score provided by two coauthors tested upon 100 sampled data, 50 for each, since it is hard to control or represent as average human performance.

\subsection{Dataset Statistics}
\label{app:data_stats}
Detailed statistics of \bench are shown in~\Cref{tab:stat}.
\begin{table}[h]
% \small
\centering
\begin{tabular}{lc}
\hline Statistics & Number \\
\hline Total Questions & 3,807\\
Total Images& 7,358\\
\hline Dev:Test& $1,901:1,906$\\
\hline
Questions with Visual Prompts & 1,946\\
Questions with Images (regions) as Choices & 2,747\\
\hline Questions with an Explanation & 300 \\
\hline Questions with Multiple Images & 2,218\\
$\quad *$ with 2 Images& 1,149\\
$\quad *$ with 3 Images& 805\\
$\quad *$ with 4 Images& 264\\
\hline
\end{tabular}
\vspace{1em}
\caption{Detailed statistics of the \bench benchmark.}
\label{tab:stat}
\end{table}

\section{Baseline Models}
\label{app:baselines}
We evaluate \bench on 16 various large multimodal LLMs. For most model families, we use the latest and best-performing available checkpoint to date. The list of baseline models are as follows:
(i) MiniGPT-4-v2~\cite{chen2023minigpt} adapts EVA~\cite{fang2023eva} as visual backbone, LLaMA2-chat (7B)~\cite{touvron2023llama} as language model backbone, and designs a linear projection layer for visual understanding abilities. 
(ii) OpenFlamingo~\cite{awadalla2023openflamingo} is an an open-source alternative to Flamingo~\cite{alayrac2022flamingo} and we use the 9B checkpoint model, built upon  CLIP~\cite{radford2021learning} vision encoder and MPT-7B language model~\cite{MosaicML2023Introducing}.
(iii - iv) InstructBLIP~\cite{dai2023instructblip} uses CLIP~\cite{radford2021learning} for vision encoder, and is fine-tuned based on BLIP-2~\cite{li2023blip} with visual instruction data. We experiment with the 7B and 13B scales, both based on the Vicuna~\cite{zheng2024judging} language model for model scaling analysis. 
(v-x) We include various LLaVa~\cite{liu2023improvedllava} models from different sources for comparison: LLaVa-internLM2-7B which is fine-tuned upon InternLM2-Chat-7B~\cite{2023internlm} language model; LLaVa-v1.5-7B-xtuner and LLaVa-v1.5-13B-xtuner that are fine-tuned upon Vicuna~\cite{zheng2024judging} from xTuner~\cite{2023xtuner}; and LLaVa-v1.5-7B, LLaVa-v1.5-13B, LLaVa-v1.6-34b from the original LLaVa papers~\cite{liu2023improvedllava,liu2024llavanext}. Compared to the v1.5 checkpoints, v1.6 checkpoint uses more reasoning, OCR, and knowledge-enhanced training data. All of the LLaVa models build upon the CLIP~\cite{radford2021learning} vision encoder.
(vi-vii) Yi-VL-6B and Yi-VL-34B\footnote{Model details can be found at \url{https://huggingface.co/01-ai/Yi-VL-6B}} are open-source models that have shown great performance on existing benchmarks. They use LLaVa structure with CLIP~\cite{radford2021learning} encoder and connect with Yi-6B-Chat or Yi-34B-Chat language models\footnote{More details are at the official website at~\url{https://www.01.ai/}}.
(viii) CogVLM~\cite{wang2023cogvlm} adds a trainable visual expert module in the attention and FFN layers to bridge different modalities better. It uses EVA-CLIP~\cite{sun2023eva} as vision encoder and Vicuna~\cite{zheng2024judging} as language backbone.
(ix)Qwen-VL~\cite{Qwen-VL} includes several powerful models that show supreme performance on existing benchmarks. We use the best model checkpoint: Qwen-VL-MAX.  
(x) GeminiProVision~\cite{team2023gemini} is one of the most powerful multimodal models, and we use the Gemini 1.0 Pro Vision version.
(xi)  Glaude 3 OPUS~\cite{claude} is a recently released multimodal model that is tested to be state-of-the-art on various datasets. We use the most powerful version: OPUS, of the Claude 3 model family.
(xii)  GPT-4~\cite{gpt4} is known to be one of the most powerful multimodal models to date. We tested on three checkpoints: GPT-4V(ision), which is gpt-4-vision-preview; GPT-4 Turbo, which is gpt-4-turbo-2024-0409; and GPT-4o, which is gpt-4o-2024-05-13. 

\noindent \textbf{GPT-4 Clarification}.
Notice that the GPT-4 performances could change if the specific checkpoint gets updated. We tested GPT-4V(ision) in March 2024, and both of GPT-4 Turbo and GPT-4o in May 2024.

\section{Analysis}
\label{app:analysis}

\begin{table}[h!]
    \centering
    % \small
    % \setlength{\tabcolsep}{1pt}
    \scalebox{0.85}{
{\fontsize{8.5pt}{10pt}\selectfont
    \begin{tabular}{lcccccccc} 
    \toprule[1.2pt]
    & \begin{tabular}{c} Validation \\$(1,901)$\end{tabular} 
    & \begin{tabular}{c} Test \\$(1,906)$\end{tabular} 
    & \begin{tabular}{c} \dreamsimshort \\$(135)$\end{tabular} 
    & \begin{tabular}{c} \countingshort \\$(120)$\end{tabular}
    & \begin{tabular}{c} \depthshort \\$(124)$\end{tabular} 
    & \begin{tabular}{c} \jigsawshort \\$(150)$\end{tabular} 
    & \begin{tabular}{c} \artshort \\$(117)$\end{tabular}
    & \begin{tabular}{c} \funcCorrshort \\$(130)$\end{tabular}    \\
    \midrule[1.2pt]
    Random Choice & 38.09& 38.09& 50& 25& 50& 50& 50 & 25\\
     Human & 95.67& 95.70& 96.70& 93.75 & 99.19 & 99.00& 95.30 & 80.77\\
    \hline \multicolumn{9}{c}{ \textbf{Open-source multimodal LLMs}} \\
    \hline 
    MiniGPT-4-v2~\cite{chen2023minigptv2} 
    & 34.23 & 34.57 & 44.44 & 13.33 & 50.81 & 34.67 & 43.59 & 20.77\\
    OpenFlamingo-v2~\cite{awadalla2023openflamingo} & 39.18 & 38.32 & 62.22 & 30.00& 54.03 & 47.33 & 52.99 & 24.62\\
    InstructBLIP-7B~\cite{dai2023instructblip} & 39.72 & 38.65 & 47.41 & 32.50& 51.61 & 52.67 & 47.01 & 23.85 \\
    InstructBLIP-13B~\cite{dai2023instructblip} & 42.24 & 39.58 & 49.63 & 30.83 & 51.61 & 52.67 & 51.28 & 29.23 \\
    LLaVA-internLM2-7B~\cite{2023internlm} & 37.71 & 36.06 & 48.89 & 55.00& 57.26 & 28.67 & 29.06 & 23.85 \\
    Yi-VL-6B~\footref{Yifootnote} & 38.72 & 41.24 & 46.67 & 55.00& 57.26 & 48.00& 39.32 & 17.69 \\
    Yi-VL-34B~\footref{Yifootnote} & 41.68 & 42.78 & 51.11 & 52.50& 50.00& 52.67 & 45.30& 31.54 \\
    LLaVA-v1.5-7B-xtuner~\cite{2023xtuner} & 39.36 & 40.81 & 47.41 & 45.83 & 51.61 & 52.67 & 47.01 & 20.00\\
    LLaVA-v1.5-13B-xtuner~\cite{2023xtuner} & 42.00 & 41.31 & 47.41 & 48.33 & 54.03 & 52.00& 47.01 & 30.00\\
    CogVLM~\cite{wang2023cogvlm} & 41.54 & 39.38 & 47.41 & 38.33 & 52.42 & 52.67 & 47.86 & 23.08 \\
    LLaVA-v1.5-7B~\cite{liu2023improvedllava} & 37.13 & 38.01 & 47.41 & 40.00& 52.42 & 11.33 & 47.01 & 20.00\\
    LLaVA-v1.5-13B~\cite{liu2023improvedllava} & 42.66 & 40.55 & 47.41 & 45.00& 53.23 & 58.00& 47.01 & 26.15 \\
    LLaVA-v1.6-34B~\cite{liu2024llavanext} & 46.80 & 45.05 & 48.89 & 66.67 & 67.74 & 54.67 & 43.59 & 20.77 \\
    \hline \multicolumn{9}{c}{ \textbf{API-based models}} \\
    \hline Qwen-VL-Max~\cite{Qwen-VL} & 40.28 & 41.94 & 51.11 & 56.67 & 58.06 & 4.67 & 38.46 & 28.46 \\ 
    Gemini Pro~\cite{team2023gemini} & 45.16 & 45.72 & 52.59 & 52.50& 40.32 & 57.33 & 50.43 & 24.62 \\
    Claude 3 OPUS~\cite{claude} & 44.05 & 44.11 & 72.59 & 50.83 & 47.58 & 32.67 & 65.81 & 21.54 \\
    GPT-4V(ision)~\cite{gpt4} & 51.14 & 51.26 & 78.52 & 60.83 & 59.68 & 70.00& 79.49 & 26.15 \\
    GPT-4 Turbo~\cite{gpt4} &  54.61 & 53.89 & 80.74 & 57.50& 66.13 & 69.33 & 79.49 & 24.62 \\
    GPT-4o~\cite{gpt4} & 60.04 & 59.03 & 72.59 & 49.17 & 74.19 & 55.33 & 82.91 & 40.77 \\
    \bottomrule
    \end{tabular}}}
    \vspace{2em}
    \setlength{\tabcolsep}{1.4pt}
    \scalebox{0.85}{
{\fontsize{8.5pt}{10pt}\selectfont
    \begin{tabular}{lcccccccc}
    \toprule[1.2pt]
    & \begin{tabular}{c} \semCorrshort \\$(139)$\end{tabular} 
    & \begin{tabular}{c} \spatialshort \\$(143)$\end{tabular}
    & \begin{tabular}{c} \localizationshort \\$(122)$\end{tabular} 
    & \begin{tabular}{c} \corrshort\\$(172)$\end{tabular} 
    & \begin{tabular}{c} \cameraposeshort \\$(133)$\end{tabular} 
    & \begin{tabular}{c} \iiwshort \\$(134)$\end{tabular} 
    & \begin{tabular}{c} \realnessshort\\$(132)$\end{tabular} 
    & \begin{tabular}{c} \iqshort\\$(150)$\end{tabular} \\
    \midrule[1.2pt]
    Random Choice & 25& 50& 50& 25& 50& 33.33&25 &25\\
     Human &96.07 & 98.25 & 98.00 & 99.42 & 92.48 & 95.14 & 100.00& 80.00\\
    \hline \multicolumn{9}{c}{ \textbf{Open-source multimodal LLMs}} \\
    \hline 
    MiniGPT-4-v2~\cite{chen2023minigptv2} & 28.78 & 44.76 & 47.54 & 26.16 & 48.87 & 30.60& 24.24 & 20.67 \\
    OpenFlamingo-v2~\cite{awadalla2023openflamingo} & 30.22 & 43.36 & 56.56 & 25.58 & 44.36 & 36.57 & 21.97 & 18.67 \\
    InstructBLIP-7B~\cite{dai2023instructblip} & 30.94 & 56.64 & 48.36 & 30.81 & 55.64 & 33.58 & 25.00& 20.00\\
    InstructBLIP-13B~\cite{dai2023instructblip} & 32.37 & 65.73 & 55.74 & 29.65 & 57.14 & 38.81 & 21.97 & 24.67 \\
    LLaVA-internLM2-7B~\cite{2023internlm} & 27.34 & 76.22 & 50.00& 27.91 & 44.36 & 32.09 & 5.30& 22.00\\
    Yi-VL-6B~\footref{Yifootnote} & 18.71 & 68.53 & 45.08 & 26.74 & 42.86 & 27.61 & 27.27 & 21.33 \\
    Yi-VL-34B~\footref{Yifootnote} & 19.42 & 71.33 & 51.64 & 26.74 & 44.36 & 38.81 & 23.48 & 24.67 \\
    LLaVA-v1.5-7B-xtuner~\cite{2023xtuner} & 28.78 & 68.53 & 36.89 & 29.07 & 38.35 & 29.85 & 36.36 & 18.67 \\
    LLaVA-v1.5-13B-xtuner~\cite{2023xtuner} & 30.94 & 69.93 & 45.08 & 29.65 & 44.36 & 38.81 & 25.76 & 24.67 \\
    CogVLM~\cite{wang2023cogvlm} & 33.09 & 63.64 & 52.46 & 29.65 & 54.14 & 29.85 & 30.30& 26.67 \\
    LLaVA-v1.5-7B~\cite{liu2023improvedllava} & 23.02 & 61.54 & 56.56 & 25.58 & 51.88 & 39.55 & 23.48 & 20.00\\
    LLaVA-v1.5-13B~\cite{liu2023improvedllava} & 32.37 & 67.83 & 52.46 & 29.07 & 44.36 & 36.57 & 31.82 & 26.00\\
    LLaVA-v1.6-34B~\cite{liu2024llavanext} & 23.74 & 74.83 & 59.02 & 30.81 & 62.41 & 31.34 & 44.70 & 26.00\\
    \hline \multicolumn{9}{c}{ \textbf{API-based models}} \\
    \hline Qwen-VL-Max~\cite{Qwen-VL} &  23.02 & 69.93 & 48.36 & 31.40& 51.88 & 36.57 & 43.94 & 21.33  \\
    Gemini Pro~\cite{team2023gemini} & 26.62 & 74.83 & 53.28 & 42.44 & 44.36 & 38.81 & 50.76 & 23.33 \\
    Claude 3 OPUS~\cite{claude} & 25.18 & 58.04 & 51.64 & 36.63 & 56.39 & 26.87 & 46.21 & 24.67 \\
     % * concatenate images & 27.86 & - & - & 23.84 & 41.35 & - & 36.36 & - \\
    GPT-4V(ision)~\cite{gpt4} & 28.78 & 72.73 & 54.92 & 33.72 & 55.64 & 38.81 & 34.09 & 22.67 \\
    GPT-4 Turbo~\cite{gpt4} & 30.94 & 69.23 & 52.46 & 52.33 & 52.63 & 32.84 & 63.64 & 32.67 \\
    GPT-4o~\cite{gpt4} & 53.96 & 69.23 & 59.84 & 75.00& 59.40& 37.31 & 79.55 & 31.33 \\
    \bottomrule[1.2pt]
    \end{tabular}}}
    \vspace{-4mm}
    \caption{\textbf{Results of different models on the \bench validation set}. The first row shows task names and number of instances. }
    \label{tab:val_results}
    \vspace{-10mm}
\end{table}
\subsection{Validation Set Results}
We include detailed scores for each task on the validation set as in Table~\ref{tab:val_results}.

\subsection{How to deal with multiple-image inputs?}
\label{app:concat_image}
\begin{table}[h!]
    \centering
    % \small
    % \setlength{\tabcolsep}{1pt}
    \scalebox{0.85}{
{\fontsize{8.5pt}{10pt}\selectfont
    \begin{tabular}{lcccccccc} 
    \toprule[1.2pt]
    & \begin{tabular}{c} \dreamsimshort \end{tabular} 
    & \begin{tabular}{c} \jigsawshort  \end{tabular} 
    & \begin{tabular}{c} \artshort \end{tabular}
    & \begin{tabular}{c} \funcCorrshort \end{tabular}
    & \begin{tabular}{c} \semCorrshort \end{tabular} 
    & \begin{tabular}{c} \corrshort \end{tabular} 
    & \begin{tabular}{c} \cameraposeshort \end{tabular} 
    & \begin{tabular}{c} \realnessshort \end{tabular} \\
    \midrule[1.2pt]
    Random Choice & 50& 50& 50 & 25 & 25& 25& 50& 25\\
     Human & 96.70 & 99.00& 95.30 & 80.77 &96.07 & 99.42 & 92.48 & 100.00\\
    \midrule[0.8pt]
    Gemini Pro~\cite{team2023gemini} & 55.88 & 54.00 & 49.57 & \textbf{32.31} & 22.14 & \textbf{37.21} & 41.35 & \textbf{45.45}\\
    * concatenate images & 42.65 & 45.33 & 48.72 & 30.77 & 27.86 & 23.84 & 41.35 & 36.36 \\
    GPT-4V(ision)~\cite{gpt4} & \textbf{83.09} & \textbf{62.67} & \textbf{78.63} & 31.54 & \textbf{30.00} & \textbf{37.21} & \textbf{58.65} & 30.30\\
    * concatenate images & 71.32 & 57.33 & 67.52 & 22.31 & 22.86 & 25.00 & 57.89 & 25.00\\
    % \hline \multicolumn{9}{c}{ \textbf{Large Language Models (LLMs): Task-agnostic Captions as Input}} \\
    % \hline GPT-4 (GPT-4V Captions) & & & & & & & &\\
    % GeminiPro (ProVision Captions) & & & & & & & &\\
    \bottomrule
    \end{tabular}}}
    \vspace{1mm}
    \caption{\textbf{Effect of concatenating multiple images on the \bench val set}.}
    \label{tab:concat_results}
    \vspace{-5mm}
\end{table}
\noindent Among all the 16 baseline models, only 2 models: GPT-4V and Gemini Pro accept multi-image inputs. Other models, especially the open-source ones, only accept single-image inputs. Since 8 out of 14 of \bench tasks require multiple images input, a natural question is, how to deal with multiple-image inputs?  
To answer this question, we convert multiple images into concatenated single image, to analyze which format would achieve better performance on multi-image understanding.
Specifically, we place the images horizontally, with a black margin in between. 
We evaluate GPT-4V and Gemini Pro with concatenated images and show results in~\Cref{tab:concat_results}.

From the experiment results, GPT-4V has shown a consistent decline in performance across all tasks when taking concatenated images as input, with the biggest decrease in \jigsaw and least decrease in \camerapose. However, the impact of concatenating images to Gemini Pro is task-dependent, with the performance decreasing in most tasks while increasing in \semCorr and remaining the same in \camerapose.

\subsection{Error analysis}
\label{app:error_analysis}

\vspace{1ex}\noindent\textbf{Open-source multimodal LLMs make similar errors.}
%different model size
%same size different model comparison
%compare different vision-encoder type: CLIP / 
%LLaVA-v1.5-7B (1187 mistakes) and LLaVA-v1.5-13B (1147 mistakes) have 899 common mistakes; they are models with same architecture and different sizes 
%LLaVA-v1.5-7B (1187 mistakes) and LLaVA-internlm2-7B (1235 mistakes) have 959 common mistakes; they are models with same size and different language models.
%LLaVA-v1.5-7B (1187 mistakes) and Yi-VL-6b (1148 mistakes) have 801 common mistakes; they are models of same size but with different language models
%LLaVA-v1.5-7B (1187 mistakes) and LLaVA-v1.5-7B-xtuner (1137 mistakes) have 938 common mistakes; they are same model with and without fine-tuning
%LLaVA-v1.5-13B (1147 mistakes) and LLaVA-v1.6-34b (1046 mistakes) have 740 common mistakes
Our comparative analysis of diverse multimodal LLMs reveals striking similarities in the cases where they fail at, highlighting that these shared mistakes are largely influenced by their vision encoder, rather than differences in model size or language model components. This is particularly apparent in the comparison between LLaVA-v1.5-7B (1187 mistakes in total) and LLaVA-v1.5-13B (1147 mistakes in total), two models of different sizes that nonetheless demonstrated 899 common mistakes. In a similar vein, when we compared LLaVA-v1.5-7B with other equal-sized models using different language model components, like LLaVA-internLM2-7B, the number of common errors remained high (959 mistakes). 
Whereas LLaVA-v1.5-7B only shares 782 and 655 common mistakes with QwenVLMax and GPT4V, respectively.
% These findings underscore the power of the vision encoder in determining the type of errors made by the multimodal LLMs, serving as an important focal point for future research and refinement to reduce these common pitfalls.

\vspace{1ex}\noindent\textbf{GPT-4V Errors:}
For each task, 10 error instances were randomly selected, and we manually analyze the total of 140 error instances sampled randomly across all tasks as follows:
Recognition failure on detailed small regions or edges (28.5\%) : the model fails to tell nuanced details, especially circles in \corr, \semCorr, \funcCorr, \depth, \iiw and boxes in \localization;
Failure to detect the location of the circled point(20\%):  the model fails to locate the circled point labeled in the images;
Failure to recognize spatial relations (14.3\%): the model fails to identify the spatial relations between left and right, or up and down;
Reasoning errors (12.9\%): while the model correctly interprets the images and the question, it fails to derive accurate reasoning for inference;
Failure to convey the overall scene impression (8.6\%): the model fails to adequately capture the general atmosphere or setting of a scene;
Rejection to answer (6.4\%): the model refuses to generate an answer;
Failure to ground or infer items mentioned in the question (5.7\%): The model is unable to locate the specific item referenced in the question within the image.

\subsection{Does self-consistency help?}
\label{app:ensemble}
To verify whether self-consistency~\cite{wang2022self} will improve the performance on \bench, we conduct five runs of GPT-4V with temperature set to 1.0. 
The self-consistency score is 48.38\%, averaged across all task, and the average single run score is 38.15\%. Note that the performance for temperature=1 decreases much compared to our default temperature=0 setting, where greedy decoding is used. From our observation, this decrease is because GPT-4V tends to face firewalls and reply ``Sorry I cannot help/assist with the question" when temperature is bigger than 0.

% Calculating the percentage improvement achieved through ensemble method yields substantial results: 'DreamSim' dataset shows an advancement of approximately 33\%, while the 'Realness' dataset has a striking increase of about 75\% in accuracy with ensemble predictions. Similarly, 'IIW' and 'ArtStyle' datasets also experienced remarkable improvements of around 79\% and 23\% respectively. Overall, the ensemble predictions significantly outperformed single predictions in all tasks, while the normalized entropy provides an inference on model uncertainty, independent from the accuracy obtained.
% The table also includes a metric of normalized entropy values, which stand as a good indicator of variance in model outputs. However, these entropy values seem to exhibit a wide range of figures, implying a rather diverse level of model output across tasks. Curiously, it appears that low entropy does not always correlate to the enhancement of performance by ensemble methods, as typified by the IIW task which has the lowest entropy but doesn't achieve a boost via ensemble method.

\section{Limitations}
\label{app:limit}
\bench makes use of data from existing image datasets, and does not cover all the visual perception abilities in the wild.
For the \realness task, we manually collected images that are publicly available from online search. 
We have made every effort to ensure that the images included in this paper are used in accordance with applicable copyright laws and are properly credited. However, if you are the copyright owner of any image included in our work and believe that its use conflicts with your licensing agreements, please contact us directly. We are committed to addressing any legitimate concerns promptly.

\end{document}